\documentclass{article}

\usepackage{arxiv}

\usepackage{algorithm}
\usepackage{algorithmic}
\usepackage{float}
\usepackage[utf8]{inputenc} 
\usepackage[T1]{fontenc}    
\usepackage{hyperref}       
\usepackage{url}            
\usepackage{booktabs}       
\usepackage{amsfonts}       
\usepackage{nicefrac}       
\usepackage{microtype}      
\usepackage{amsmath}
\usepackage{lipsum}
\usepackage{graphicx}
\usepackage{nth}

\usepackage{graphicx}
\usepackage[table]{xcolor}

\usepackage{subcaption}

\begin{document}
\title{Deep Ensemble for Rotorcraft Attitude Prediction}

\author{
Hikmat Khan \\
\textit{Dept. of Electrical and }\\ \textit{Computer Engineering}\\
\textit{Rowan University}\\
Glassboro, New Jersey, USA \\
bouaynaya@rowan.edu
\And
Nidhal C. Bouaynaya\\
\textit{Dept. of Electrical and }\\ 
\textit{Computer Engineering} \\
\textit{Rowan University}\\
Glassboro, New Jersey, USA \\
bouaynaya@rowan.edu
\And
Ghulam Rasool\\
\textit{Dept. of Machine Learning} \\
\textit{Moffitt Cancer Center}\\
Tampa, Florida, USA \\
ghulam.rasool@moffitt.org\\
\And
\textbf{Tyler Travis} \\
Federal Aviation Administration \\
Atlantic City, New Jersey, \\
United States\\
\And
\textbf{Lacey Thompson} \\
Federal Aviation Administration \\
Atlantic City, New Jersey, \\
United States\\
\And
\textbf{Charles C. Johnson}\\
Research Engineer, Flight Test Engineer, and Program Lead,\\
Federal Aviation Administration \\
Egg Harbor Township,New Jersey, \\
}
\maketitle    
\begin{abstract}
Historically, the rotorcraft community has experienced a higher fatal accident rate than other aviation segments, including commercial and general aviation. To date, traditional methods applied to reduce incident rates have not proven hugely successful for the rotorcraft community. Recent advancements in artificial intelligence (AI) and the application of these technologies in different areas of our lives are both intriguing and encouraging. When developed appropriately for the aviation domain, AI techniques may provide an opportunity to help design systems that can address rotorcraft safety challenges. Our recent work demonstrated that AI algorithms could use video data from onboard cameras and correctly identify different flight parameters from cockpit gauges, e.g., indicated airspeed. These AI-based techniques provide a potentially cost-effective solution, especially for small helicopter operators, to record the flight state information and perform post-flight analyses. We also showed that carefully designed and trained AI systems can accurately predict rotorcraft attitude (i.e., pitch and yaw) from outside scenes (images or video data). Ordinary off-the-shelf video cameras were installed inside the rotorcraft cockpit to record the outside scene, including the horizon. The AI algorithm was able to correctly identify rotorcraft attitude at an accuracy in the range of 80\%. In this work, we combined five different onboard camera viewpoints to improve attitude prediction accuracy to 94\%.
Our current approach, which is referred to as ensembled prediction, significantly increased the reliability in the predicted attitude (i.e., pitch and yaw). For example, in some camera views, the horizon may be obstructed or not visible. The proposed ensemble method can combine visual details recorded from other cameras and predict the attitude with high reliability. In our setup, the five onboard camera views included pilot windshield, co-pilot windshield, pilot Electronic Flight Instrument System (EFIS) display, co-pilot EFIS display, and the attitude indicator gauge. Using video data from each camera view, we trained a variety of convolutional neural networks (CNNs), which achieved prediction accuracy in the range of 79\% to 90\%. We subsequently ensembled the learned knowledge from all CNNs and achieved an ensembled accuracy of 93.3\%. Our efforts could potentially provide a cost-effective means to supplement traditional Flight Data Recorders (FDR), a technology that to date has been challenging to incorporate into the fleets of most rotorcraft operators due to cost and resource constraints. Such cost-effective solutions can gradually increase the rotorcraft community's participation in various safety programs, enhancing safety and opening up helicopter flight data monitoring (HFDM) to historically underrepresented segments of the vertical flight community.


\end{abstract}

\section{Introduction}
As the premier agency for promoting and ensuring aviation safety, the Federal Aviation Administration (FAA) continually strives to improve safety. The FAA recognizes the importance of participating in Helicopter Flight Data Monitoring (HFDM) programs and encourages their increased utilization to improve flight safety and operational efficiency. Indeed, rotorcraft safety was one of the agency's top ten most wanted list of safety improvements in 2017-2018 and continues to be a high priority still in 2021. Organizations including the FAA, National Transportation Safety Board (NTSB), and the United States Helicopter Safety Team (USHST) are strong proponents of flight data recorders (FDRs). These organizations and other industry partners are working together to promote helicopter flight data monitoring (HFDM) programs as one possible mitigation strategy to reduce the rotorcraft fatal accident rate. However, despite all of these efforts by various safety organizations, barriers to widespread implementation of FDRs and adaptation of HFDM still exist. These include, but are not limited to, the technical skills required to certify, install, and maintain an FDR, skilled resources who can perform HFDM analyses, and the costs associated with the acquisition, certification, and installation of the FDR. Traditional FDRs require a Supplemental Type Certificate (STC) or Field Approval (FA) to install and operate under the Rotorcraft Flight Manual (RFM). On average, the initial acquisition cost of an FDR can range from \$5,000 - \$50,000. Given a range of factors, rotorcraft, in general, have a lower participation rate in FDM programs than other forms of aviation, including commercial fixed-wing or part 121 air carriers.

\begin{figure*}[htb]
\centering
 \hfill 
  \subfloat[Pilot windshield]{
	\begin{minipage}
	{
	   0.30\textwidth}
	   \centering
	   \includegraphics[width=1\textwidth]{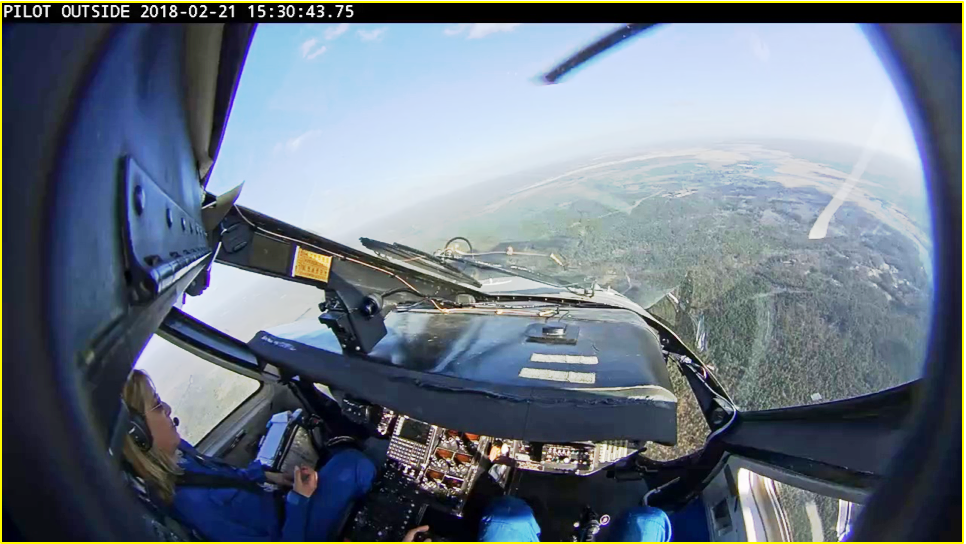}
	\end{minipage}
	}
 \hfill 	
  \subfloat[Co-pilot windshield]{
	\begin{minipage}
	{
	   0.30\textwidth}
	   \centering
	   \includegraphics[width=1\textwidth]{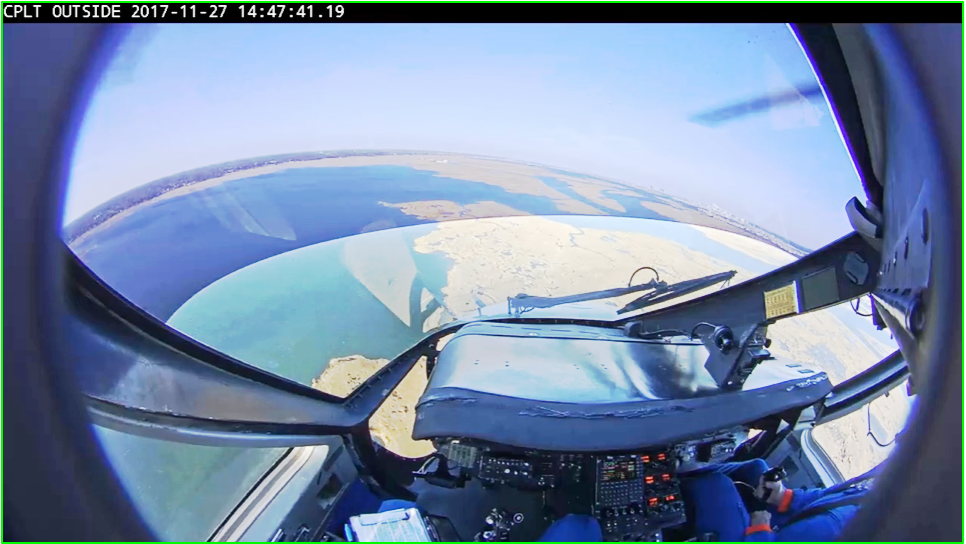}
	\end{minipage}
	}
 \hfill 
  \subfloat[Pilot broom closet]{
	\begin{minipage}
	{
	   0.30\textwidth}
	   \centering
	   \includegraphics[width=1\textwidth]{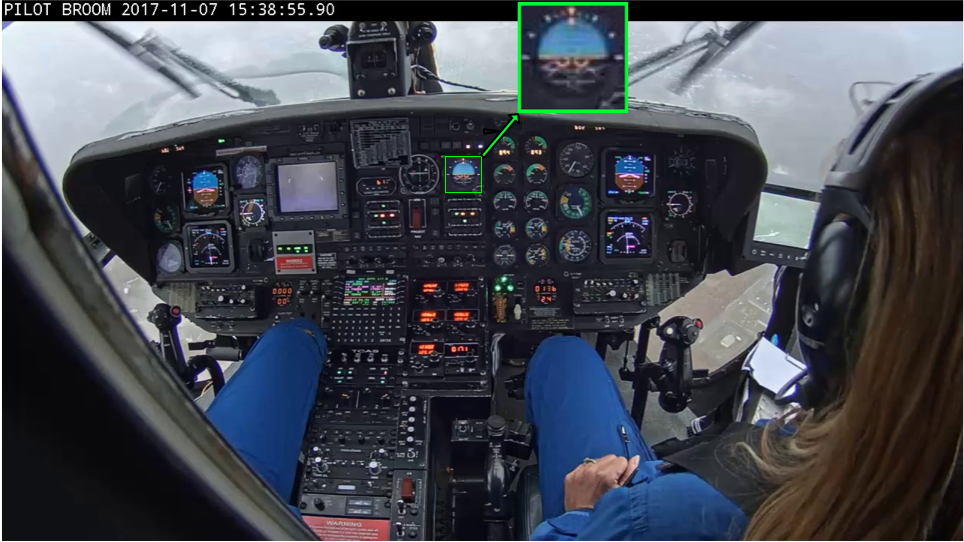}
	\end{minipage}
	}
 \vfill
 \hfill	
  \subfloat[Pilot EFIS display]{
	\begin{minipage}
	{
	   0.48\textwidth}
	   \centering
	   \includegraphics[width=1.5in]{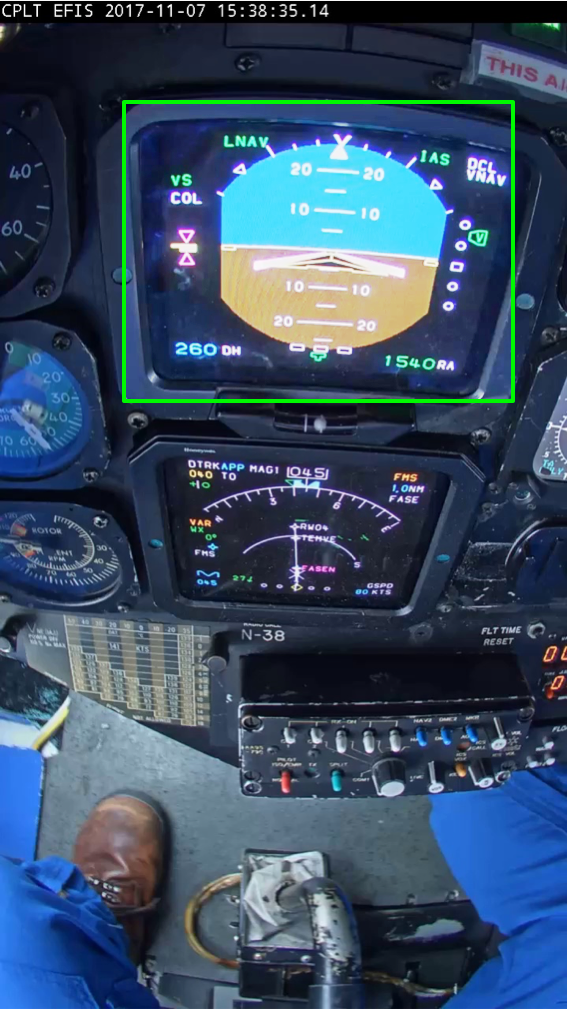}
	\end{minipage}
	}
 \hfill	
  \subfloat[Co-pilot EFIS display]{
	\begin{minipage}
	{
	   0.48\textwidth}
	   \centering
	   \includegraphics[width=1.45in]{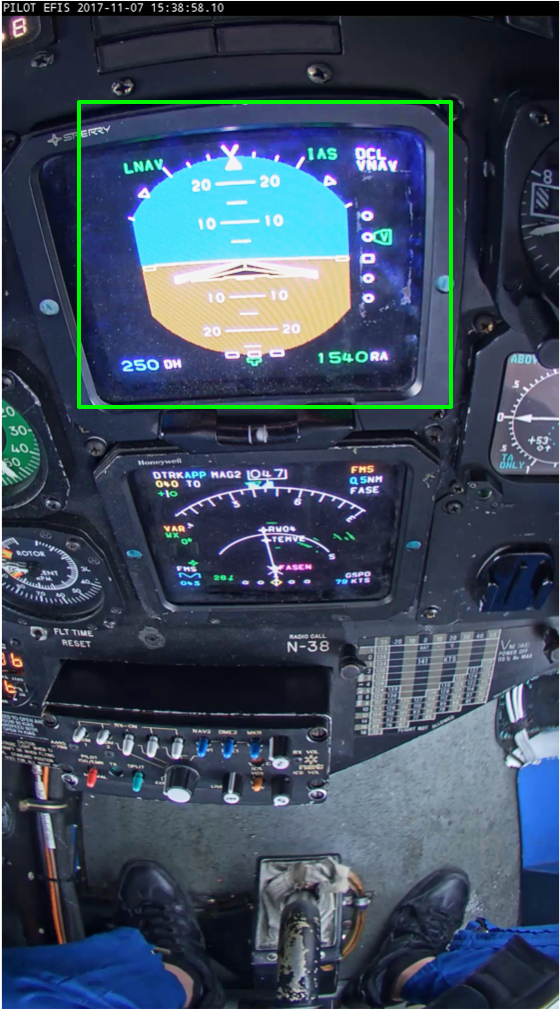}
	\end{minipage}
	}
 \hfill

\caption{A set of representative images from five different cameras mounted inside the helicopter cockpit are presented. In the cases of (a) and (b), we used the whole video frame as an input for CNNs. While in the case of (c), (d), and (e), the recorded image was cropped. The input to the CNN included the areas inside the highlighted green rectangles. (best viewed in color)}
\label{fig:sample_of_views}
\end{figure*}

\begin{figure*}[htb]
\centering
 \hfill 
  \subfloat[Blurred vision]{
	\begin{minipage}
	{
	   0.323\textwidth}
	   \centering
	   \includegraphics[width=1\textwidth]{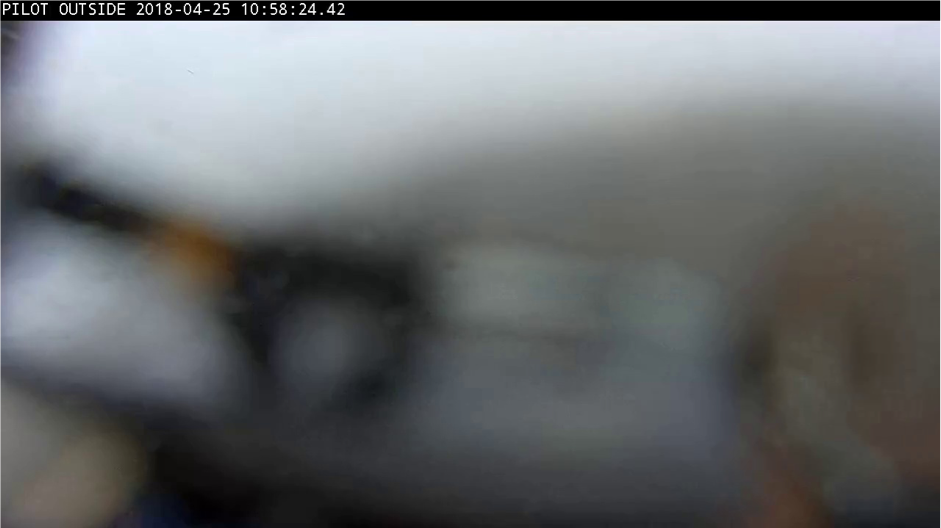}
	\end{minipage}
	}
 \hfill 	
  \subfloat[Rainy day]{
	\begin{minipage}
	{
	   0.32\textwidth}
	   \centering
	   \includegraphics[width=1\textwidth]{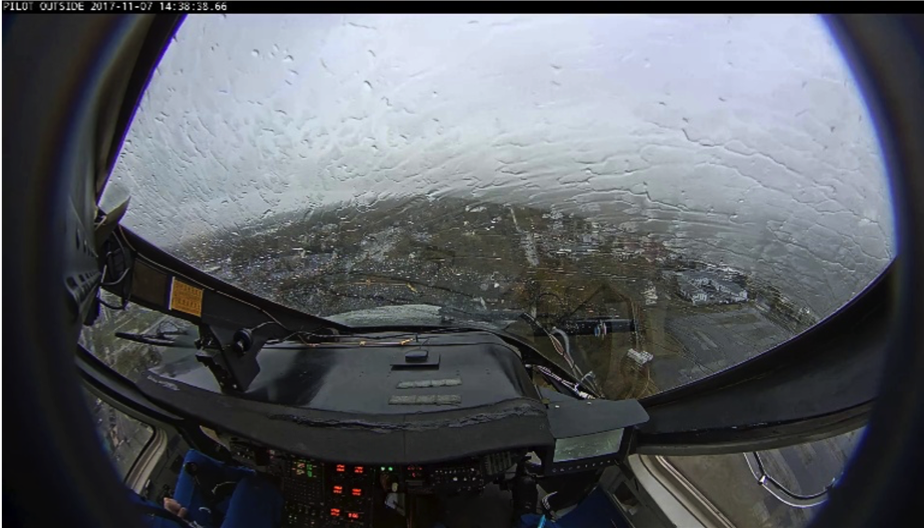}
	\end{minipage}
	}
 \hfill 
  \subfloat[Night flight]{
	\begin{minipage}
	{
	   0.323\textwidth}
	   \centering
	   \includegraphics[width=1\textwidth]{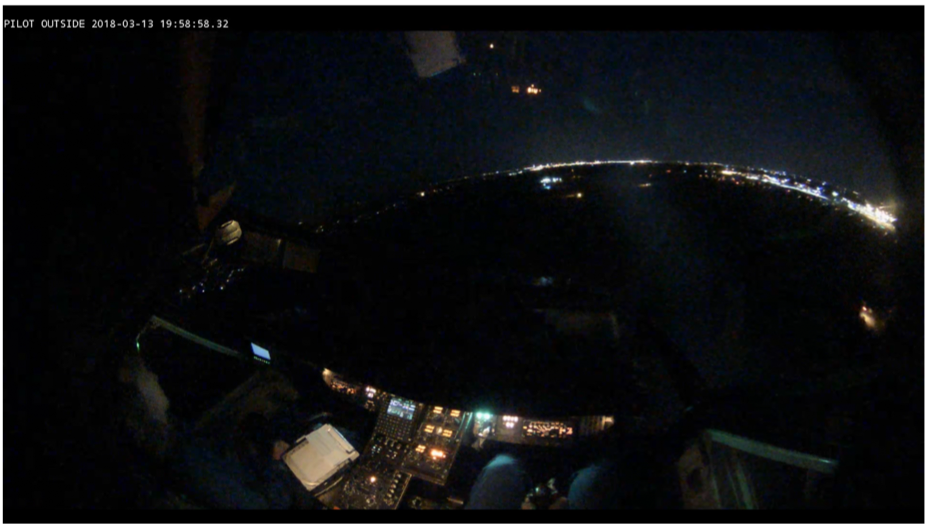}
	\end{minipage}
	}
\vfill
 \hfill	
  \subfloat[Glare]{
	\begin{minipage}
	{
	   0.35\textwidth}
	   \centering
	   \includegraphics[width=1\textwidth]{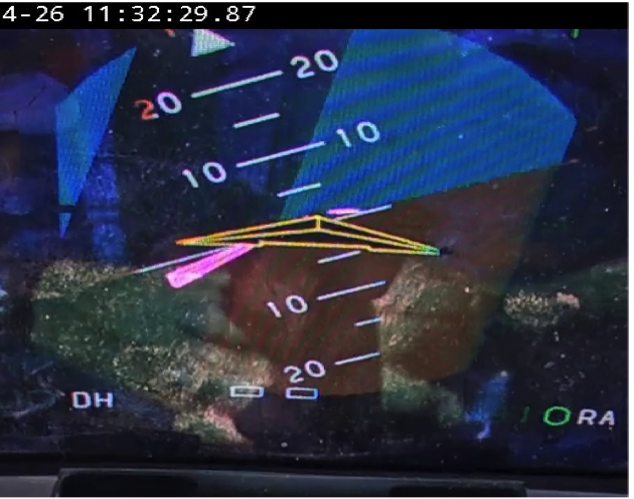}
	\end{minipage}
	}
 \hfill	
  \subfloat[EFIS display]{
	\begin{minipage}
	{
	   0.30\textwidth}
	   \centering
	   \includegraphics[width=1\textwidth]{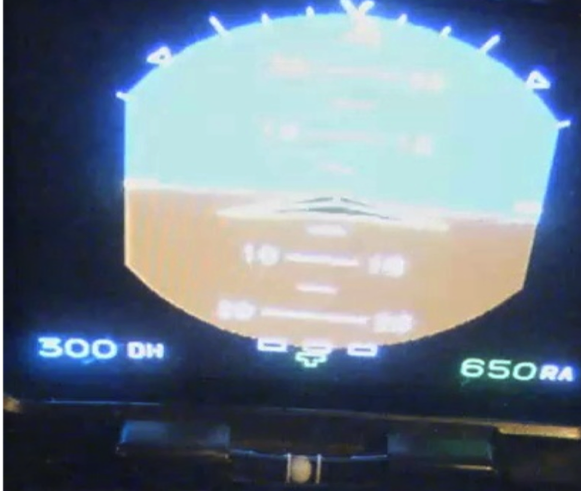}
	\end{minipage}
	}
 \hfill	
   \subfloat[Varying ambient light]{
	\begin{minipage}
	{
	   0.30\textwidth}
	   \centering
	   \includegraphics[width=1\textwidth]{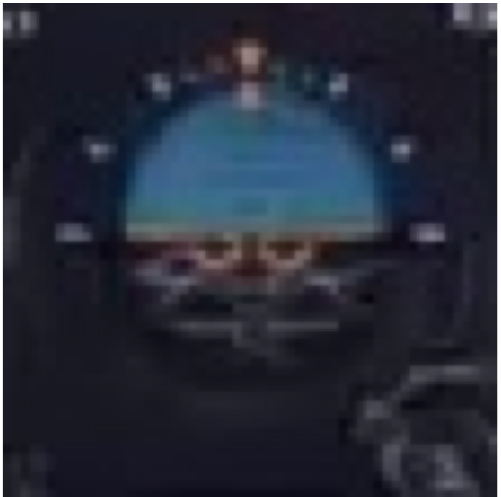}
	\end{minipage}
	}
 \hfill

\caption{Representative images from different camera views. These images show the prediction of attitude from one gauge can be challenging, and an ensemble approach may benefit due to the availability of different types of data related to the attitude of the rotorcraft.(best viewed in color)}
\label{fig:challenges}
\end{figure*}

{Inexpensive and off-the-shelf video cameras mounted inside the cockpit may offer a potential alternative to traditional FDRs. Even small helicopter operators often have access to or have the financial means to purchase one or more off-the-shelf video cameras. These cameras can potentially record all the data that traditional FDRs record. Moreover, onboard cameras may provide supplementary data that may not be available in some FDRs. The recorded video data from onboard cameras can be used for various analyses and inquiries. The examples including estimation of flight parameters from instrument panel gauges, flight replay during post-accident investigations, estimation of rotorcraft attitude, and any other visual information that can be extracted from video data.}

In this work, we recorded video data from five different onboard cameras for the accurate prediction of rotorcraft attitude. The cameras included the pilot windshield, co-pilot windshield, pilot Electronic Flight Instrument System (EFIS) display, co-pilot EFIS display, and the attitude indicator gauge. Figure \ref{fig:sample_of_views} shows representative images from all onboard cameras. We used in-flight videos from these cameras to build training datasets for our AI models. We used the video datasets to train a variety of convolutional neural networks (CNNs). Later, we combined the knowledge learned by all CNNs using the ensemble approach, which improved the attitude prediction accuracy to 93.3\%. Our proposed ensemble approach improves attitude prediction accuracy especially when the horizon curve is obstructed or not visible due to bad weather or other conditions. Our results from this work and previous publications support the viability of an inexpensive onboard camera-based solution for flight parameters' estimation and attitude prediction. Such cameras-based solutions do not require any special modification to the helicopter's avionics, communications, or display systems and are more suitable for legacy helicopters. Importantly, the solution's cost-effective nature will encourage the rotorcraft community to participate in these voluntary safety programs.

The paper is organized as follows. The \textit{Related Work} section presents a brief description of the current AI-based approaches proposed to increase rotorcraft safety. In \textit{Methodology} section, we explain the data acquisition methodology and describe our experimental setup. \textit{Results} section presents experimental results along with a discussion on our results. Finally, we conclude the current research in \textit{Conclusion} section.

\section{Related Work}
Various AI models based on deep neural networks have obtained above human-level performance on many different tasks. Examples in the computer vision domain include image classification, object detection, and image segmentation \cite{Bauer2013ASO, Srinivas2016ATO, Zhao2018ObjectDW}. Computer vision tasks are primarily approached using the well-known type of artificial neural networks, i.e., convolutional neural network or CNN. A great deal of research has recently focused on developing various kinds of CNNs, including EfficientNet, VGG-16, VGG-19, ResNet, Inception, and Xception \cite{EfficientNet, Simonyan, He2016DeepRL, Chollet2016XceptionDL}. These CNNs are also referred to as deep neural networks as they are built using tens or hundreds of layers of artificial neurons. CNNs have demonstrated the ability to learn complicated features directly from the image or video data in an increasingly complex hierarchy.

Deep neural networks and CNNs are being proposed to tackle various challenging tasks in the aviation community \cite{Hikmat, Hikmat_Forum_76, Alligier2015MachineLA, Alligier}. Khan \emph{et al.} showed that CNNs could infer different flight parameters from video data recorded during flights \cite{Hikmat}. The authors showed that their trained models could accurately estimate various flight parameters, including airspeed and engine torque, directly from the instrument panel videos, with the core purpose to facilitate the post-flight analysis. 

Alligier \emph{et al.} used machine learning techniques to improve airspeed estimation during aircraft climbing \cite{Alligier2015MachineLA}. In another work, the same authors used machine learning to estimate the mass of ground-based aircraft during climb prediction \cite{Alligier}. Kenneth \emph{et al.} applied a natural language processing (NLP) technique of structural topic modeling to the Aviation Safety Reporting System (ASRS) corpus \cite{Kenneth}. The authors identified subjects, patterns, and areas that needed further analysis in the ASRS corpus \cite{Kenneth}. Gianazza \emph{et al.} trained a variety of machine learning algorithms to predict the workload of air traffic controllers \cite{Gianazza}. The closest research work to ours was performed by Shin \emph{et al.} \cite{Shin}. The author proposed a conventional computer vision strategy using hand-engineered features to predict rotorcraft attitude from onboard cameras \cite{Shin}. The authors manually analyzed each video frame and marked the natural horizon line. Later, the classical machine learning approach of DBSCAN Clustering was employed to estimate the roll and bank angles \cite{Shin}. However, the proposed method was computationally expensive and lacked scalability for the inference of multiple parameters in large video datasets.

\begin{figure*}[htb]
\centering
 \hfill 
  \subfloat[Pitch Angle (deg)]{
	\begin{minipage}
	{
	   0.45\textwidth}
	   \centering
	   \includegraphics[width=1\textwidth]{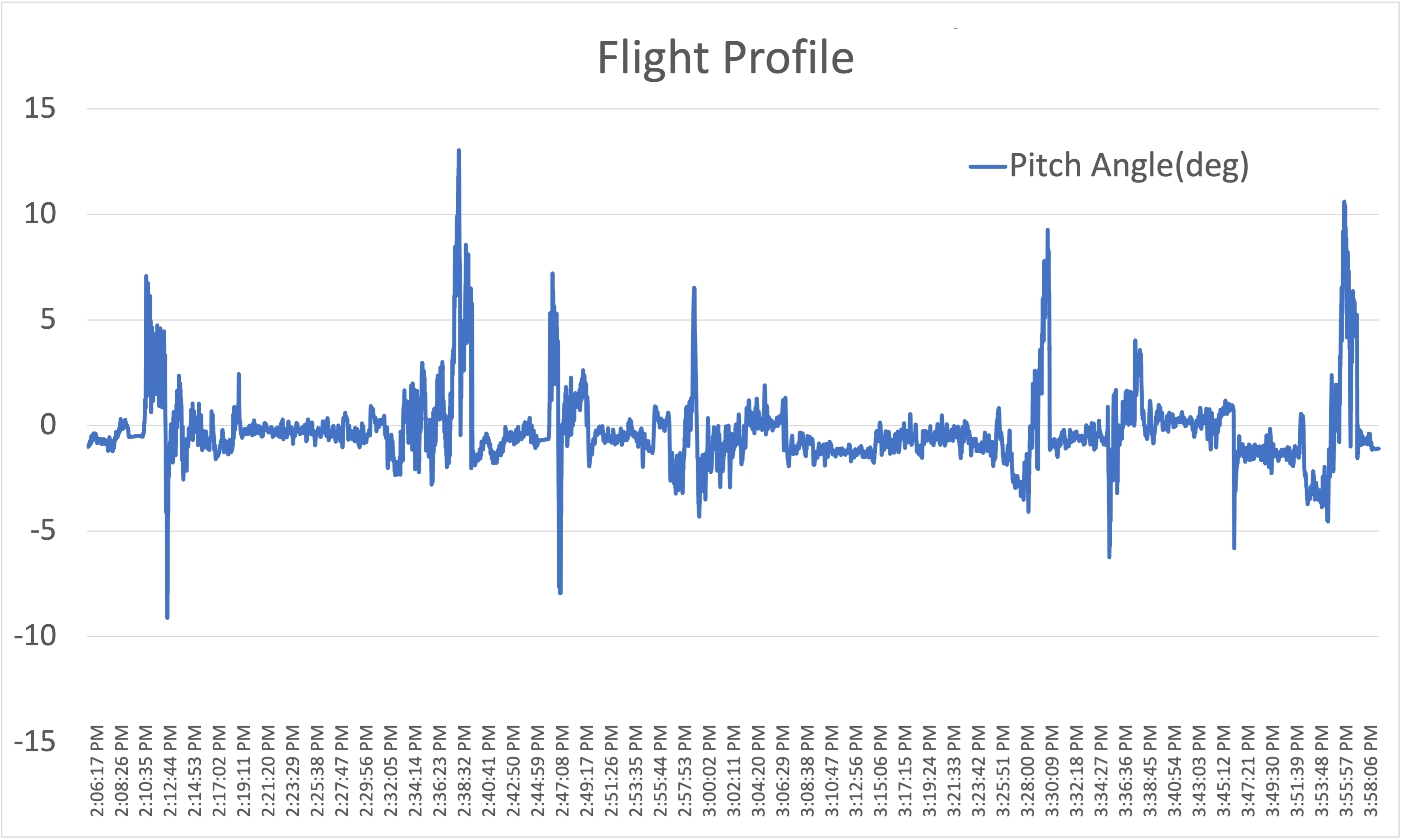}
	\end{minipage}
	}
 \hfill 	
  \subfloat[Roll Angle (deg)]{
	\begin{minipage}
	{
	   0.45\textwidth}
	   \centering
	   \includegraphics[width=1\textwidth]{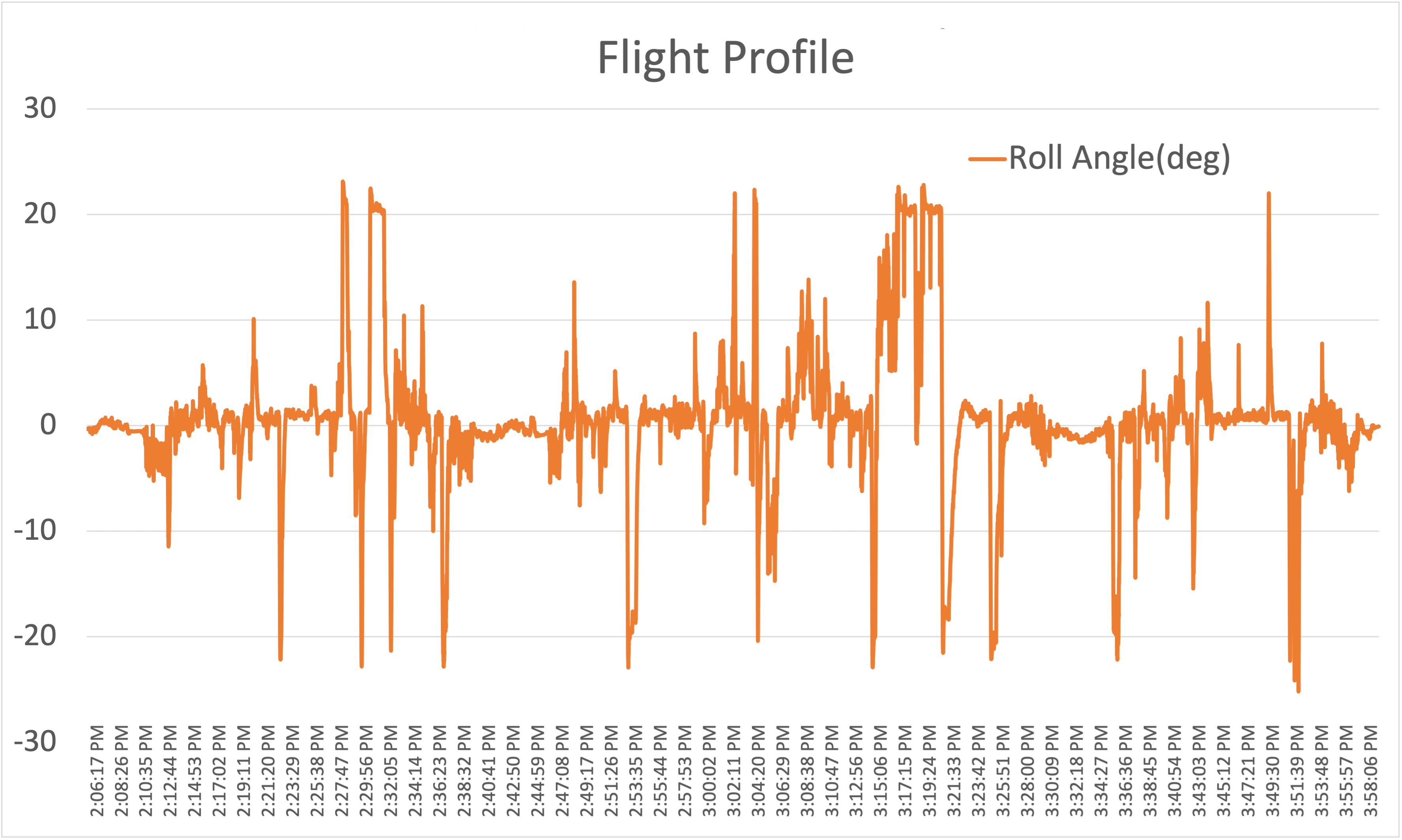}
	\end{minipage}
	}
 \hfill 
\caption{The pitch and roll values for the whole duration of the flight are presented in figure (a) and (b), respectively. The y-axis shows pitch values in degrees and x-axis represents flight duration. (best viewed in color)}
\label{fig:flight_roll}
\end{figure*}

\section{Methodology}

We trained multiple different deep learning models for each dataset to predict rotorcraft attitudes (i.e., pitch and yaw). Each dataset was build using the video stream from a different camera. The five cameras installed in the cockpit collected different details on the horizon curve from their own viewpoints. For instance, the pilot and co-pilot windshield view directly record the horizon curve, as seen through the windshield. The pilot and co-pilot EFIS display and the attitude indicator gauge on the instrument penal present a different visual representation of the same horizon curve. Figure \ref{fig:sample_of_views} shows the sample images of each of the five onboard cameras . 

\subsection{Convolutional Neural Networks (CNNs)}
CNNs are inspired by the human visual system and have obtained state-of-the-art performance in various computer vision tasks \cite{Shen, Chen,Redmon}. CNNs can learn features from input images in a hierarchical fashion taking advantage of the spatial coherence in images, and do not require domain-specific knowledge, i.e., feature engineering. The early layers of a CNN learn standard features, such as edges or lines, while subsequent layers learn complex domain-dependant features. The convolutional layers of a CNN perform convolution operations between the input image and learnable parameters (also referred to as filters or kernels). A CNN may have multiple layers of convolutional kernels depending upon the complexity of the task to be learned. The features extracted by a convolutional layer are passed through a nonlinear function, such as rectified linear unit (ReLU). The nonlinear functions are also referred to as activation functions or simply activations. The activation functions are generally followed by a max-pooling layer, which reduces the dimensionality of the input. At the end of the last convolutional layer, the extracted features are flattened and densely connected to the next layer, referred to as the fully-connected layer. Finally, a softmax function is used to produce class scores or probabilities.  

\subsubsection{Weight Initialization:}
In deep neural networks, appropriate weights (or parameters) initialization can reduce the convergence time and computational cost. In our experiments, we initialized our weights with ImageNet \cite{He2016DeepRL}. 
 
\subsubsection{Activation Functions:}
The nonlinear activation functions are generally introduced into every layer of a neural network. Some commonly used activation functions include sigmoid, hyperbolic tangent (tanh), ReLU, Leaky ReLU (LeakyReLU), and parametric ReLU \cite{ian_goodfellow}. We used ReLU activation function in all of our experiments.

\subsubsection{Max-Pooling Layers:}
Pooling is used to reduce the dimensionality of the input features by sub-sampling. Pooling operation makes the CNN invariant to small intensity and illumination changes as well as translations. The commonly used pooling operations include max-pooling and average-pooling \cite{ian_goodfellow}. The max-pooling operation select features with the maximum value in the pooling region. The average pooling calculates the average of the features in the pooling region. We used the max-pooling operation in all experiments.

\subsubsection{Regularization:}
A common problem with deep neural networks is overfitting, i.e., lack of generalization to the unseen data. Several regularization schemes have been proposed to avoid overfitting, i.e., $L_1$-Regularization, $L_2$-Regularization, batch normalization, and dropout \cite{ian_goodfellow}. Batch normalization is performed by calculating parameterized mean and standard deviation of the input and processed data at various layers of the CNN during the training phase. Dropout is a commonly used and effective technique for regularization \cite{dropout} where a randomly selected set of neurons are tuned off (forced to zero) at each forward pass. Dropout forced each neuron to learn and contribute independently to the overall output of the CNN. We used both batch normalization and dropout techniques in our experiments.

\subsubsection{Fully Connected Layers and the Softmax Function:}
The output of the convolutional layers is called the feature map. The feature map is flattened (i.e., vectorized) and densely connected to the next layer, a fully connected layer. A CNN may have multiple fully connected layers. The output of the last fully connected layer is referred to as class scores and becomes an input to the softmax function. Softmax function nonlinearly normalizes input class scores to numbers between 0 and 1. The highest score is considered the classification decision, i.e., the predicted class label by the CNN.
 
\subsubsection{Loss Function:}
The loss function measures the error between the predicted class label by the CNN and the ground truth label. The training process of a neural network is aimed to minimize the loss function by optimally adjusting weights or parameters. We used the categorical cross-entropy loss function in all our experiments.

\subsection{Data Acquisition:} We mounted five onboard cameras with various viewpoints inside the cockpit to record the instrument panel and the horizon inside an S-76 helicopter. The pilot and co-pilot perspectives of the horizon were captured with pilot and co-pilot windshield cameras mounted above the pilot and co-pilot. Similarly, the pilot and co-pilot perspectives of the EFIS displays were recorded using the two separate cameras. The fifth camera continuously recorded the different gauges on the instrument panel. The representative images can be seen in Figure \ref{fig:sample_of_views}. Table \ref{tbl:flight_summary} presents the total duration of the available videos for each camera view. The rotorcraft was equipped with an onboard Helicopter Flight Data Recorder (HFDR). Both HFDR and cameras (i.e., each frame of flight video) were timestamped using a time server. The timestamps were used to annotate individual frames of flight videos with the corresponding HFDR recordings. 

\begin{table}[ht] \begin{minipage}{\columnwidth} \centering
\caption{Total flight duration for five different datasets, i.e., camera views. The flight duration is in the standard time format (i.e. hh:mm:ss).}
\label{tbl:flight_summary}
\arrayrulecolor{green}
\begin{tabular}{lcc} 
\hline \hline
Camera veiw & Flight duration  \\ \hline
Pilot Windshield & 44:37:10 \\  
Co-pilot Windshield & 55:38:34 \\  
Pilot EFIS  & 23:42:35 \\  
Co-pilot EFIS & 34:09:50 \\  
Artificial Attitude Indicator  & 15:20:00 \\

 \hline \hline
\end{tabular} \end{minipage}
\end{table}

\begin{table}[ht] \begin{minipage}{\columnwidth} \centering
\caption{Definition of classes for attitude. We used a threshold $\alpha = 3$ and defined 9 discrete classes. Abbreviations used: NU - nose up, ND - nose down, RP - roll positive; RN - roll negative, and  L - level or steady-state.}
\label{tbl:classes_defination}
\arrayrulecolor{green}
\begin{tabular}{lccc} 
\hline \hline
Class & Description & Pitch(P) & Roll(R) \\ \hline

0 & NU & P $>$ $\alpha$ & $-\alpha \leq$ R $\leq +\alpha$\\
1 & ND & P $<-\alpha$ &$-\alpha \leq$ R $\leq +\alpha$\\
2 & RR & $-\alpha \leq$ P $\leq +\alpha$ & R $> \alpha$\\
3 & RL & $-\alpha \leq$ P $\leq +\alpha$ &R $< -\alpha$\\
4 & NU \& RP & P $> \alpha$ & R $> \alpha$\\
5 & NU \& RN & P $> \alpha$ & R  $< -\alpha$\\
6 & ND \& RP & P $< -\alpha$ & R $> \alpha$\\
7 & ND \& RN & P $< -\alpha$ & R $< -\alpha$\\
8 & L & $-\alpha \leq$ P $\leq +\alpha$ & $-\alpha \leq$ R $\leq +\alpha$ \\ \hline \hline
\end{tabular} \end{minipage}
\end{table}

\subsection{Classes Definition:} The HFDR recorded reading for the attitude (i.e., pitch and yaw) are real numbers. Figure \ref{fig:flight_roll} present the pitch and roll attitude values for one of the flights of S-76 helicopter, respectively. We define nine classes for the attitude. The nine classes are: class 0 - nose down (ND), class 1 - nose up (NU), class 2 - roll positive (RP), class 3 - roll negative (RN), class 4 - ND and RP, class 5 - NU and RP, class 6 - ND and RN, class 7 - NU and RN, and class 8 - level and steady-state (L). Table \ref{tbl:classes_defination} presents definition of nine classes. The $\alpha$ takes the user-defined values and defines the boundary among nine classes. In our experiments we set the value as $\alpha = 3$.

\subsection{Experimental Setup:} We trained four CNNs for each video dataset. CNN architectures that we considered for attitude prediction included EfficientNet, VGG16, VGG19, ResNet50, and Xception and InceptionV3 \cite{EfficientNet, Hu2017SqueezeandExcitationN, Simonyan, He2016DeepRL, Chollet2016XceptionDL, Chen2019HybridTC}. The models were trained on the training set and then evaluated on a separate test set. In all experiments, we initialized the models with ImageNet weights and fine-tuned them to our dataset \cite{keras_models}. All experiments were performed using the Adam optimizer with a consistent batch size of 256 \cite{adam}. The remaining parameters of the Adam optimizer were initialized to default values, as discussed in \cite{adam}.

\section{Results}

\begin{table*}
\centering
\caption{Average prediction accuracies of all models trained on five camera views. The last two rows present accuracies for two ensemble models. We used majority voting strategy for creating ensemble models.}
\label{tbl:average_results}
\arrayrulecolor{green}
\begin{tabular}{lcccccc}  \hline \hline
Model & Pilot & Co-pilot & Pilot & Co-pilot & Attitude & Ensemble\\
 & windshield & windshield & EFIS & EFIS & Gauge & Performance  \\\hline

EfficientNetB0 & 86\% & 83.8\% & 86\% &86\% &76.67\%&-\\ 
ResNet50 & 77.1\% & 81.7\%  & 85\% &-&-&-\\ 
VGG16 & 85\% & \bf{89.1}\%  & \bf{91.6}\% &\bf{90}\%&74.4\%&-\\ 
VGG19 & \bf{87}\% & 88\%  & 84\% &-&\bf{79.1}\%&-\\ 
InceptionV3 & - & -  & - &85.6\%&-&-\\ 
Xception & - & -  & - &88.1\%&76.8\%&-\\ 
\hline  
Deep Ensemble (excluding attitude gauge models) & - & -  & - & - & - &\bf{92.5\%}\\ 
Deep Ensemble (including all 20 models) & - & -  & - & - & - &\bf{93.3\%}\\ 

 \hline \hline
\end{tabular} 
\end{table*}

In Table \ref{tbl:average_results}, we present average accuracy values for all models trained on five different datasets. Each row of the table represents a different CNN, and each column represents a dataset (i.e., camera view). The last column shows the performance of two ensemble approaches. The ensemble approaches combine the attitude predictions of all trained models (i.e., 20 models) on five camera views using a majority voting strategy. The first ensemble approach did not include models trained with the attitude gauge datasets. The second approach included all the models. 

As evident from Table \ref{tbl:average_results}, the proposed two ensemble approaches obtained higher averaged accuracy (i.e., 92.5\% and 93.3\%) compared to other individual models trained on different camera views. The improvement in the predictive performance using the ensemble approach supports our hypothesis that combining data from multiple cameras will improve prediction accuracy.

\begin{figure*}[htb]
\centering
 \hfill 
  \subfloat[VGG16]{
	\begin{minipage}
	{
	   0.45\textwidth}
	   \centering
	   \includegraphics[width=1\textwidth]{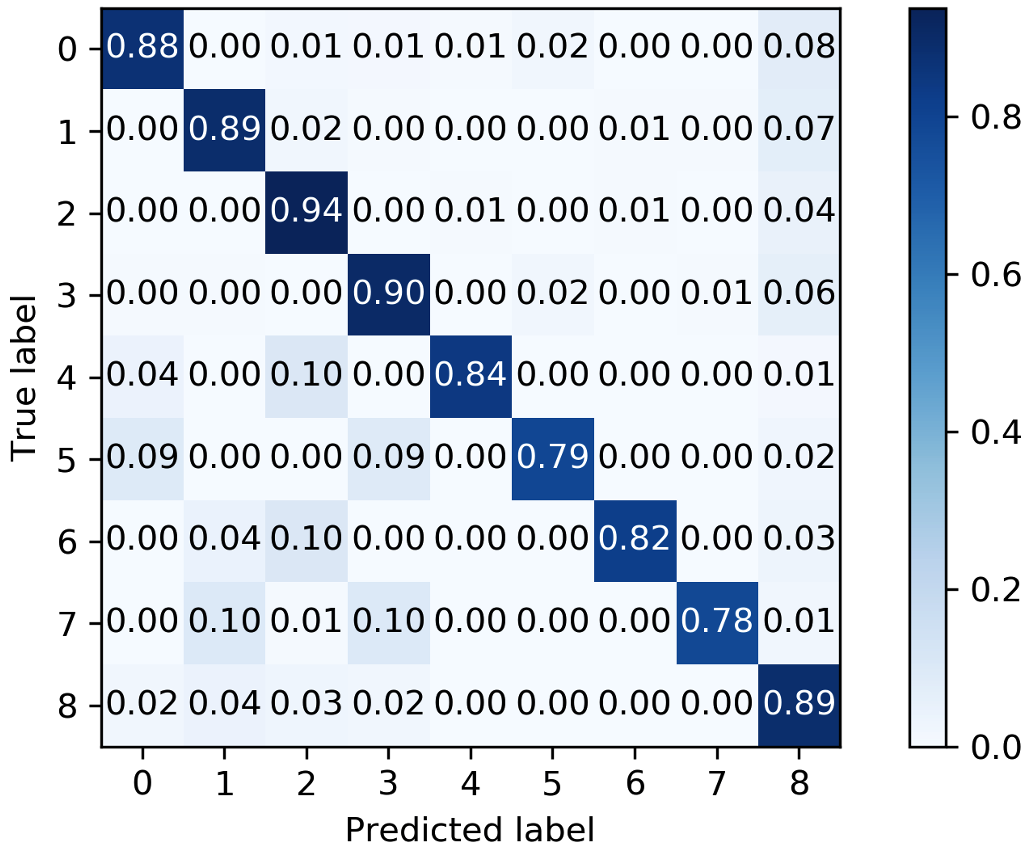}
	\end{minipage}
	}
 \hfill 	
  \subfloat[VGG19]{
	\begin{minipage}
	{
	   0.45\textwidth}
	   \centering
	   \includegraphics[width=1\textwidth]{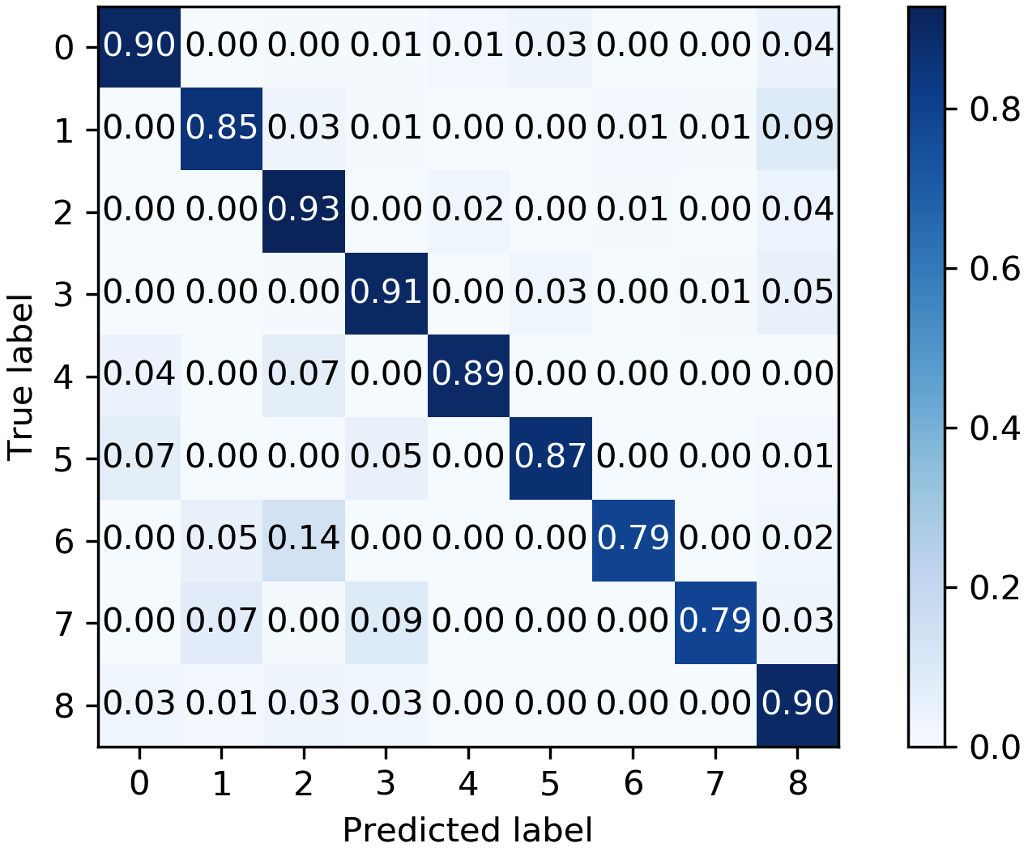}
	\end{minipage}
	}
 \hfill 
 \vfill
  \hfill 	
  \subfloat[EfficientNet]{
	\begin{minipage}
	{
	   0.45\textwidth}
	   \centering
	   \includegraphics[width=1\textwidth]{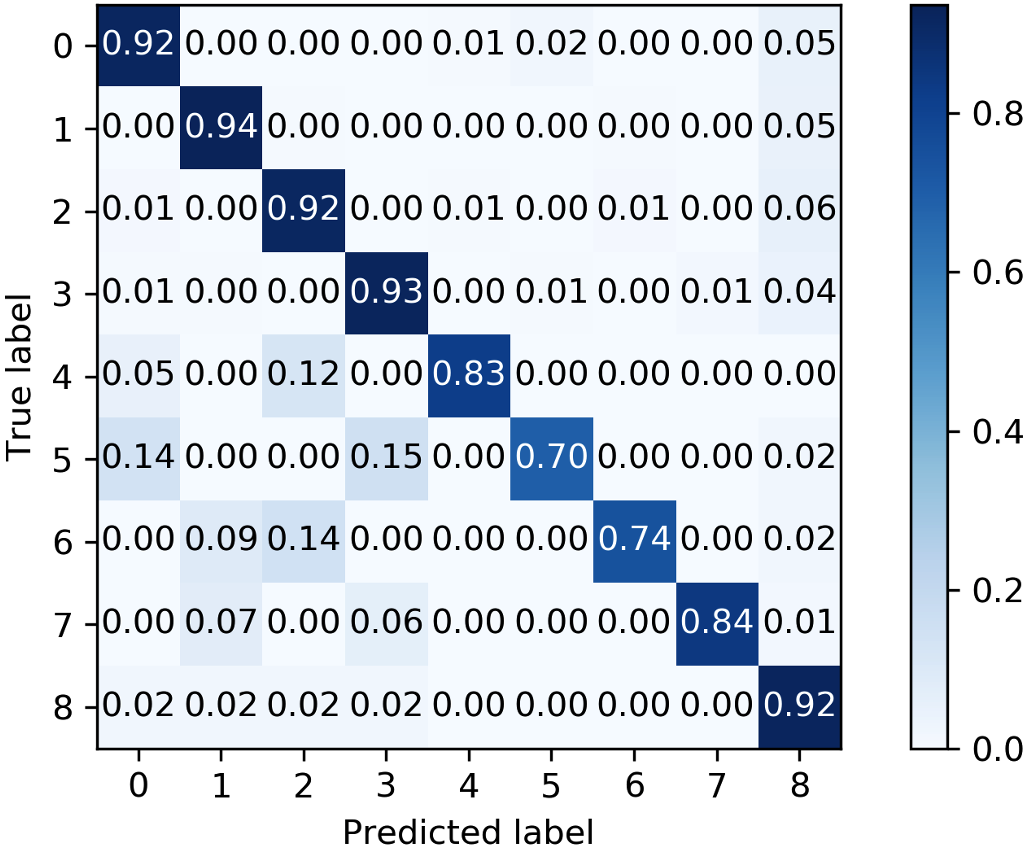}
	\end{minipage}
	}
 \hfill 	
  \subfloat[ResNet50]{
	\begin{minipage}
	{
	   0.45\textwidth}
	   \centering
	   \includegraphics[width=1\textwidth]{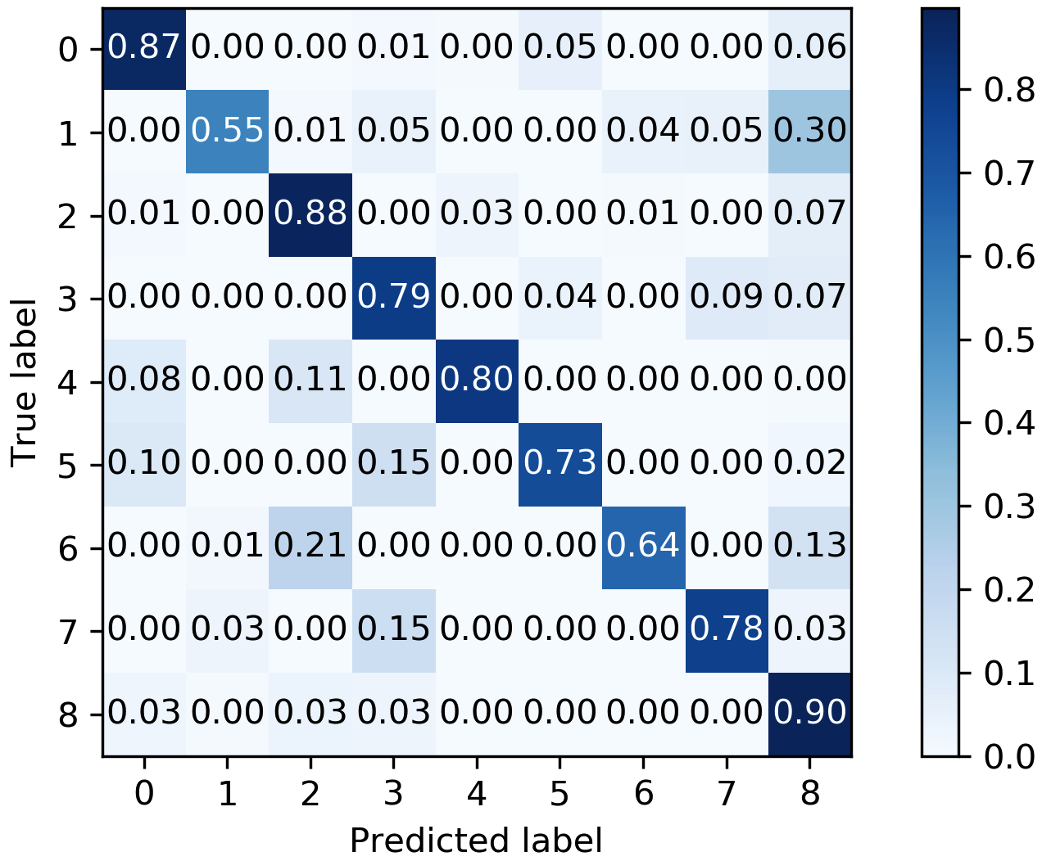}
	\end{minipage}
	}
 \hfill 
\caption{Class-wise normalized confusion matrices for 4 CNNs (i.e. VGG16, VGG19, EfficientNet, and ResNet50) used for attitude prediction using pilot windshield camera view are presented.}
\label{fig:pilot_ws_cm_1}
\end{figure*}

\begin{figure*}[htb]
\centering
 \hfill 
  \subfloat[EfficientNet]{
	\begin{minipage}
	{
	   0.45\textwidth}
	   \centering
	   \includegraphics[width=1\textwidth]{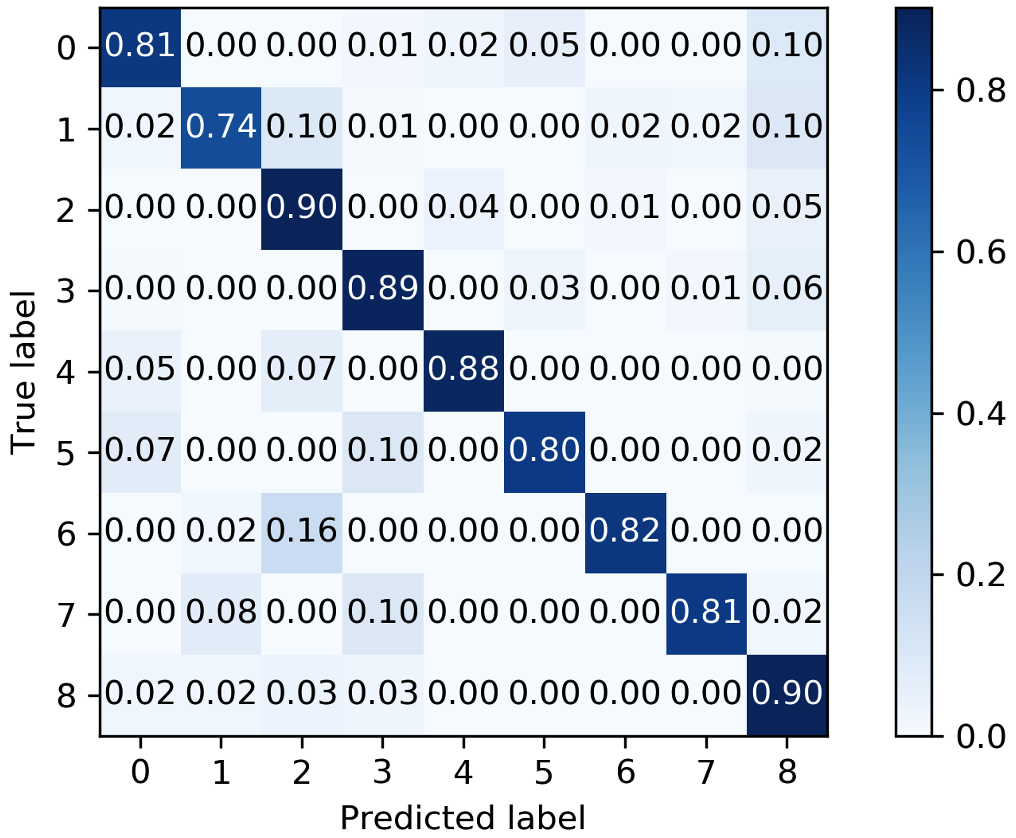}
	\end{minipage}
	}
 \hfill 	
  \subfloat[VGG19]{
	\begin{minipage}
	{
	   0.45\textwidth}
	   \centering
	   \includegraphics[width=1\textwidth]{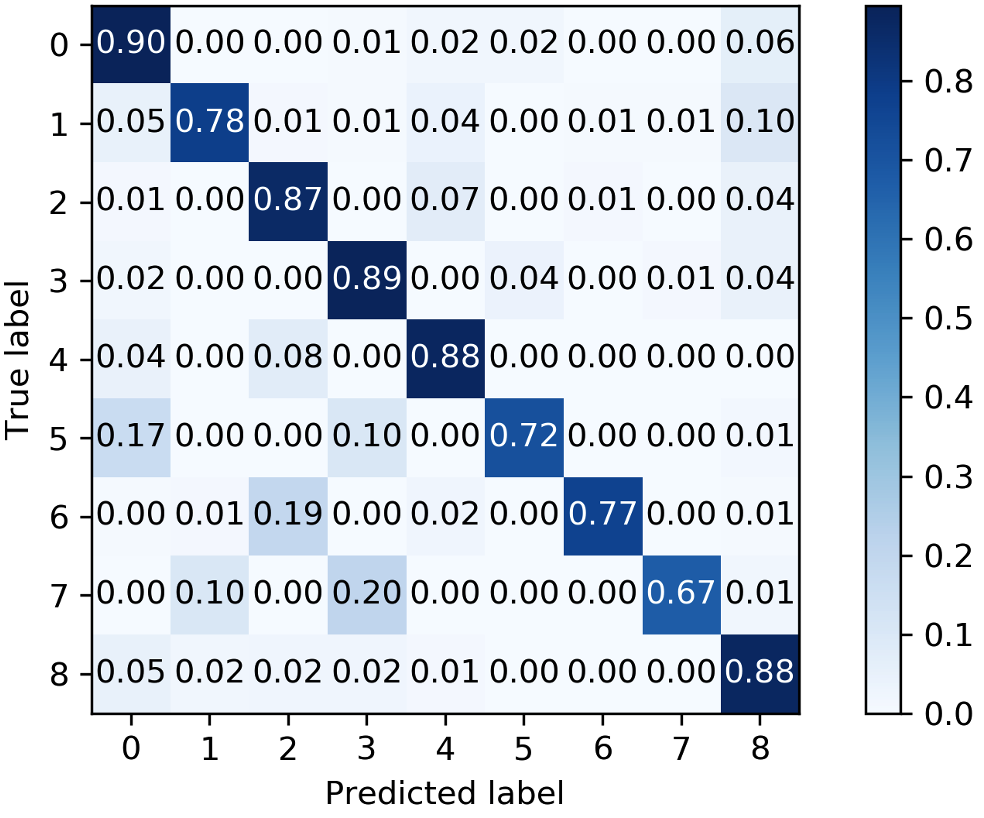}
	\end{minipage}
	}
 \hfill 
 \vfill
  \hfill 	
  \subfloat[EfficientNet]{
	\begin{minipage}
	{
	   0.45\textwidth}
	   \centering
	   \includegraphics[width=1\textwidth]{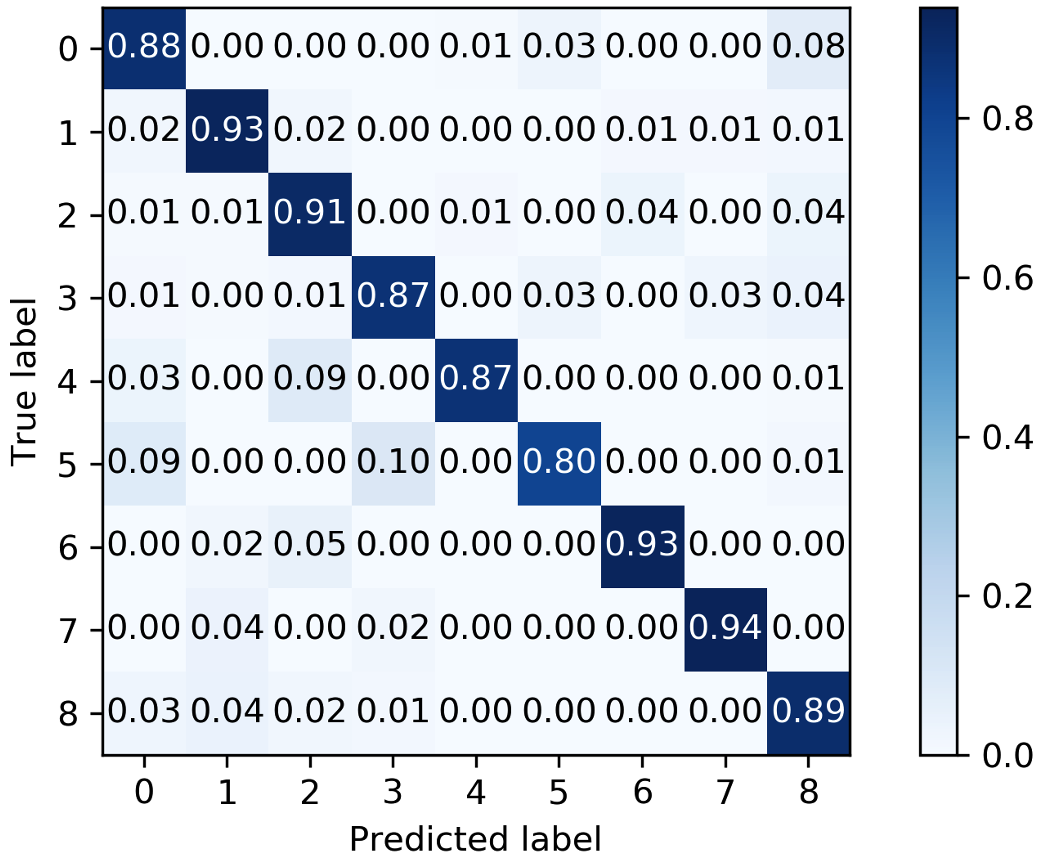}
	\end{minipage}
	}
 \hfill 	
  \subfloat[ResNet50]{
	\begin{minipage}
	{
	   0.45\textwidth}
	   \centering
	   \includegraphics[width=1\textwidth]{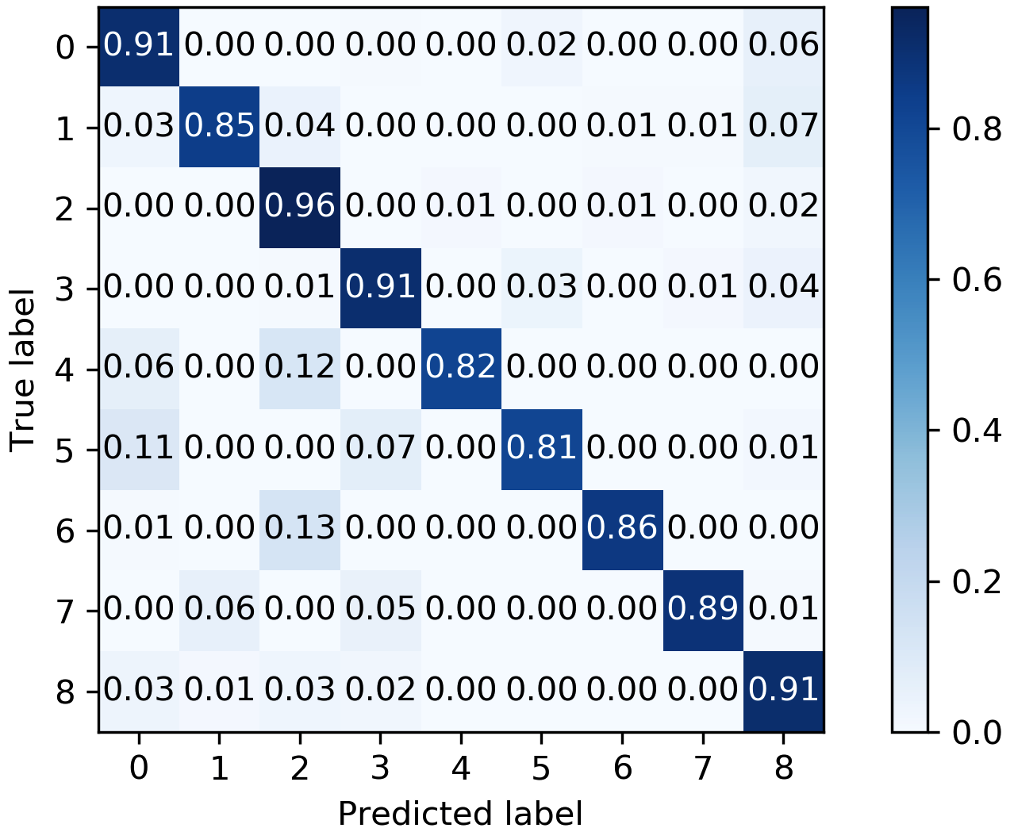}
	\end{minipage}
	}
 \hfill 
\caption{Class-wise normalized confusion matrices for 4 CNNs (i.e. EfficientNet, ResNet50, VGG16, and VGG19) used for attitude prediction using co-pilot windshield camera view are presented.}
\label{fig:co_pilot_ws_cm_1}
\end{figure*}

\begin{figure*}[htb]
\centering
 \hfill 
  \subfloat[EfficientNet]{
	\begin{minipage}
	{
	   0.45\textwidth}
	   \centering
	   \includegraphics[width=1\textwidth]{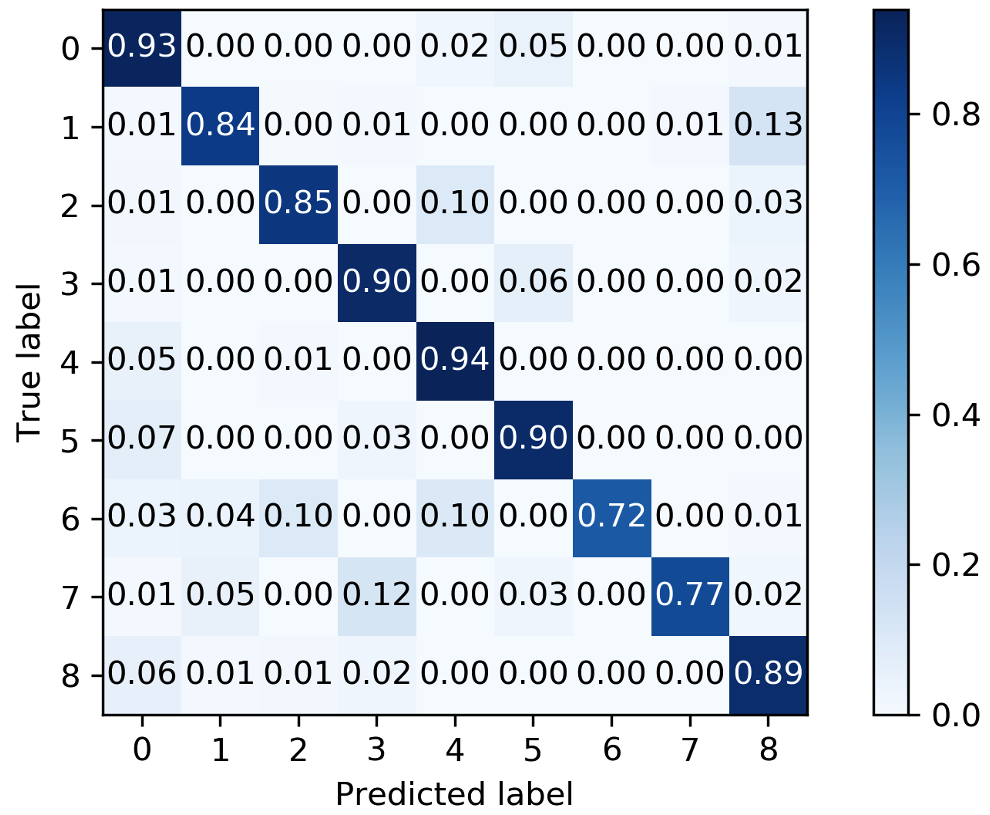}
	\end{minipage}
	}
 \hfill 	
  \subfloat[ResNet50]{
	\begin{minipage}
	{
	   0.45\textwidth}
	   \centering
	   \includegraphics[width=1\textwidth]{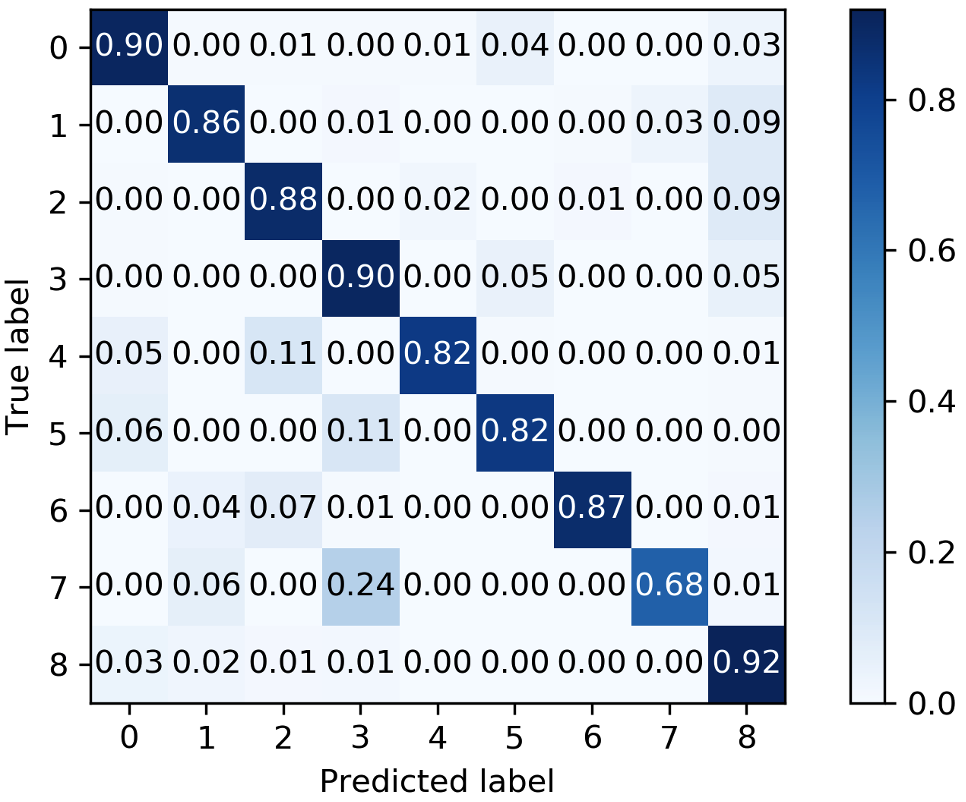}
	\end{minipage}
	}
 \hfill 
 \vfill
  \hfill 	
  \subfloat[VGG16]{
	\begin{minipage}
	{
	   0.45\textwidth}
	   \centering
	   \includegraphics[width=1\textwidth]{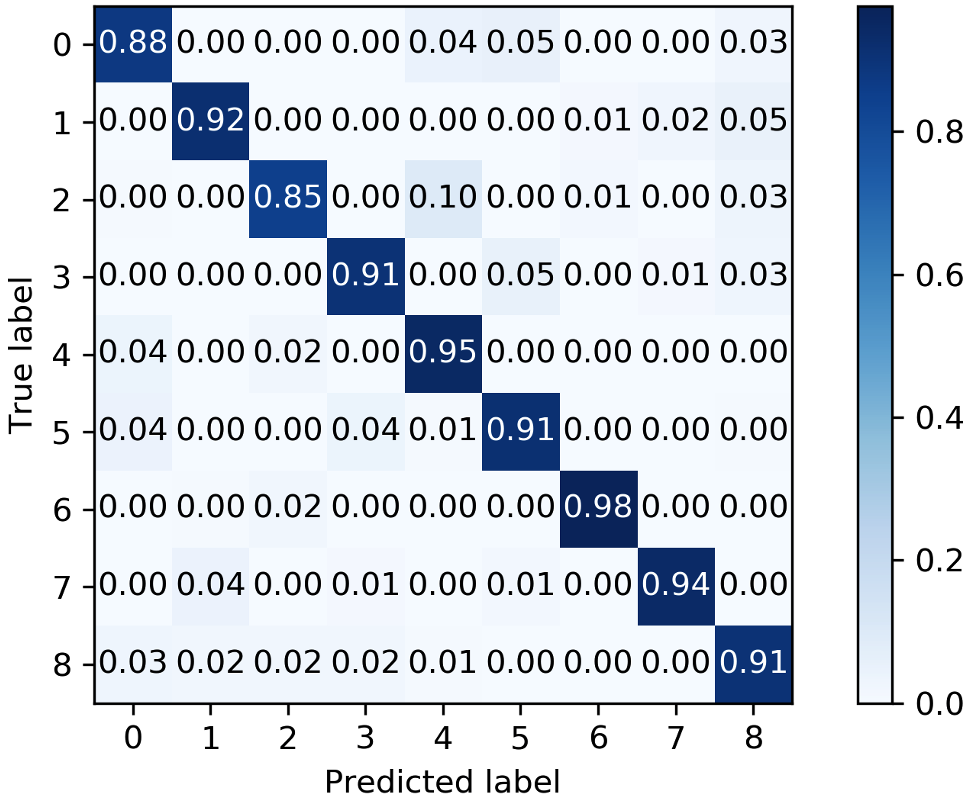}
	\end{minipage}
	}
 \hfill 	
  \subfloat[VGG19]{
	\begin{minipage}
	{
	   0.45\textwidth}
	   \centering
	   \includegraphics[width=1\textwidth]{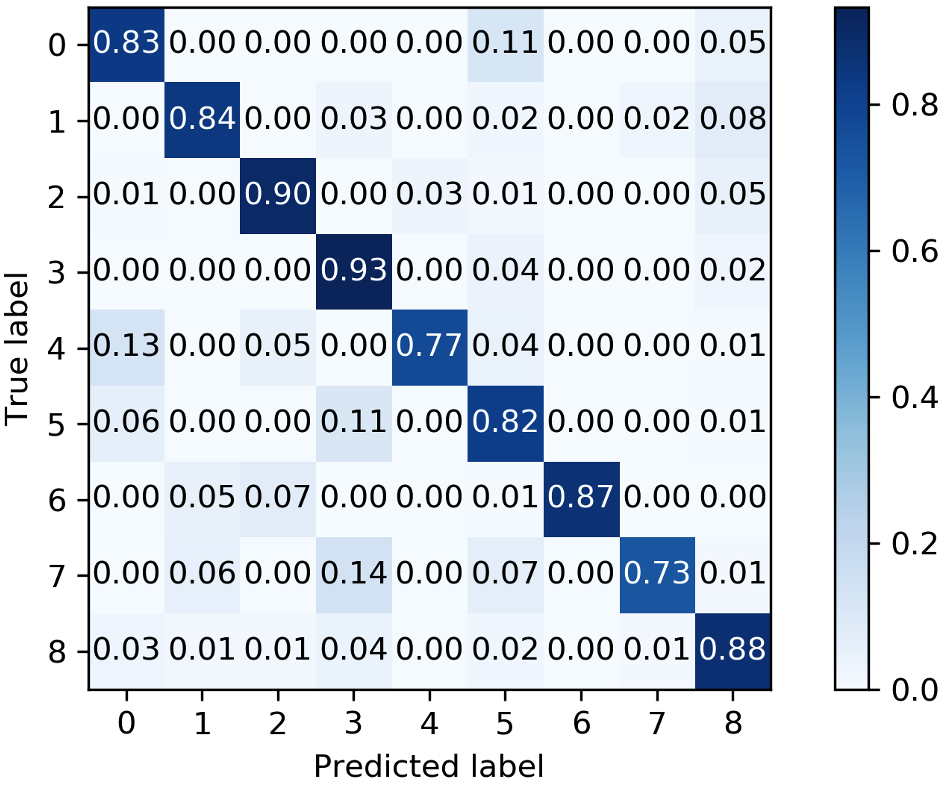}
	\end{minipage}
	}
 \hfill 
\caption{Class-wise normalized confusion matrices for 4 CNN architectures (i.e. EfficientNet, RestNet50, VGG16 and VGG19) used for attitude prediction using pilot EFIS display are presented.}
\label{fig:pilot_efis_cm_1}
\end{figure*}

\begin{figure*}[htb]
\centering
 \hfill 
  \subfloat[EfficientNet]{
	\begin{minipage}
	{
	   0.45\textwidth}
	   \centering
	   \includegraphics[width=1\textwidth]{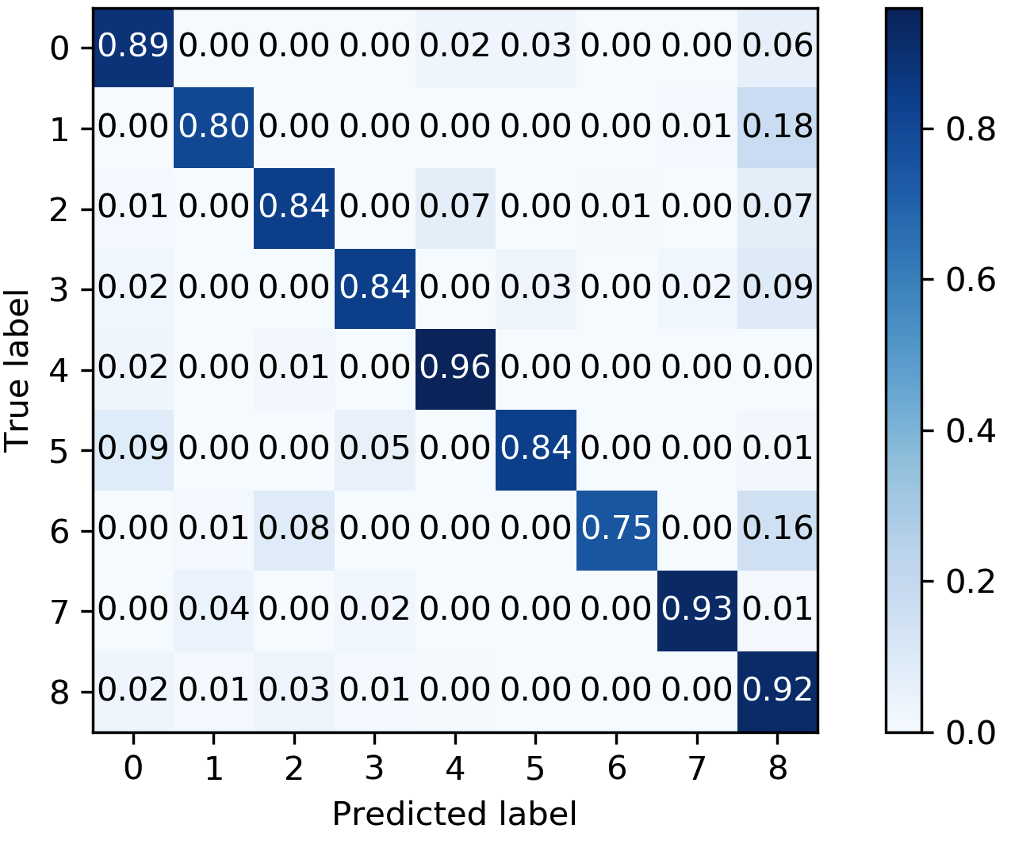}
	\end{minipage}
	}
 \hfill 	
  \subfloat[Xception]{
	\begin{minipage}
	{
	   0.45\textwidth}
	   \centering
	   \includegraphics[width=1\textwidth]{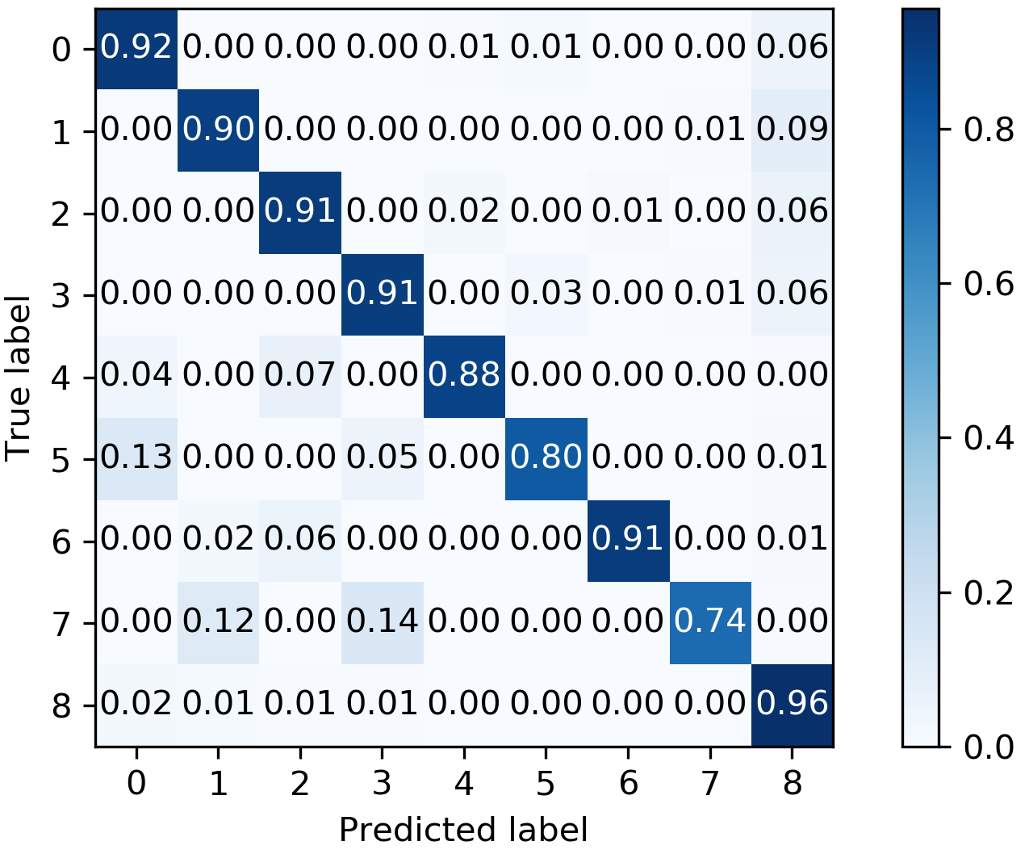}
	\end{minipage}
	}
 \hfill 
 \vfill
  \hfill 	
  \subfloat[VGG16]{
	\begin{minipage}
	{
	   0.45\textwidth}
	   \centering
	   \includegraphics[width=1\textwidth]{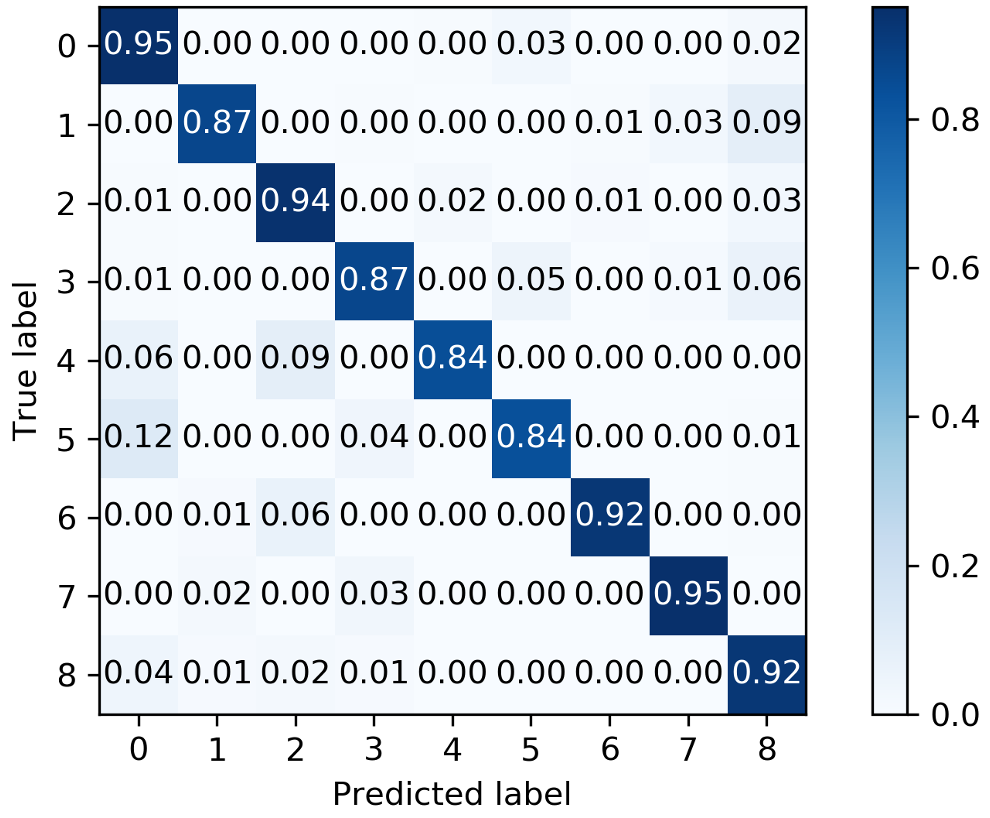}
	\end{minipage}
	}
 \hfill 	
  \subfloat[InceptionV3]{
	\begin{minipage}
	{
	   0.45\textwidth}
	   \centering
	   \includegraphics[width=1\textwidth]{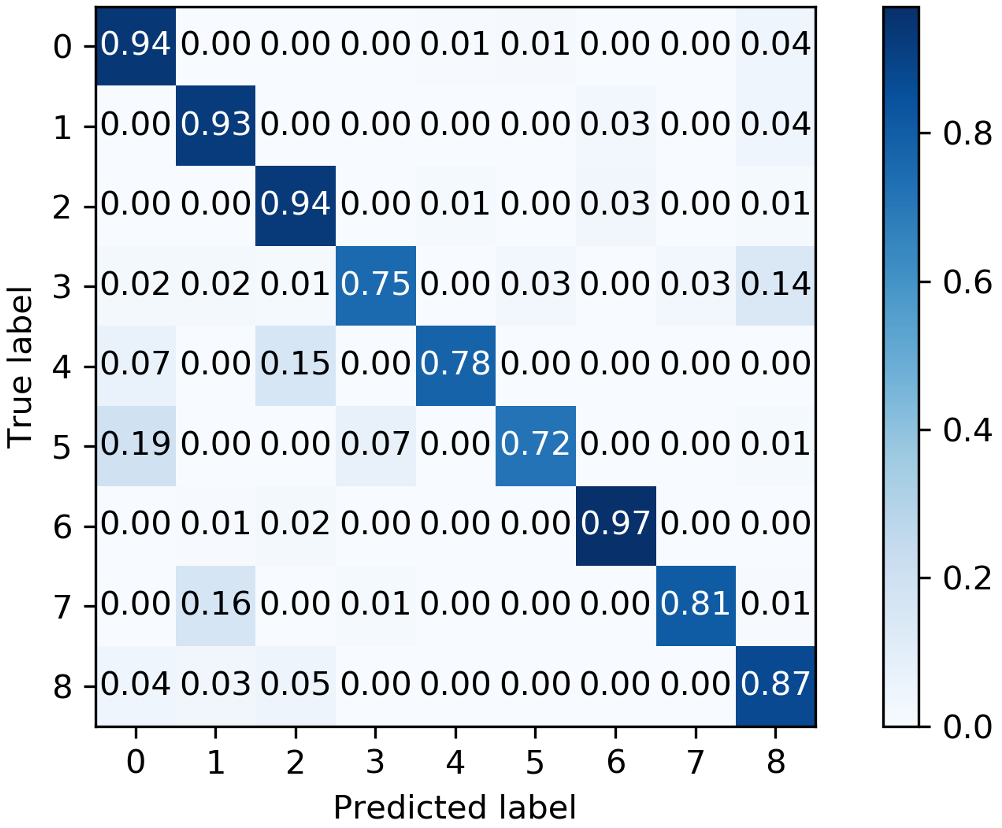}
	\end{minipage}
	}
 \hfill 
\caption{Class-wise normalized confusion matrices for 4 CNN architectures (i.e. EfficientNet, Xception, VGG16, and InceptionV3) used for attitude prediction using co-pilot EFIS display are presented.}
\label{fig:co_pilot_efis_cm_1}
\end{figure*}

\begin{figure*}[htb]
\centering
 \hfill 
  \subfloat[EfficientNet]{
	\begin{minipage}
	{
	   0.45\textwidth}
	   \centering
	   \includegraphics[width=1\textwidth]{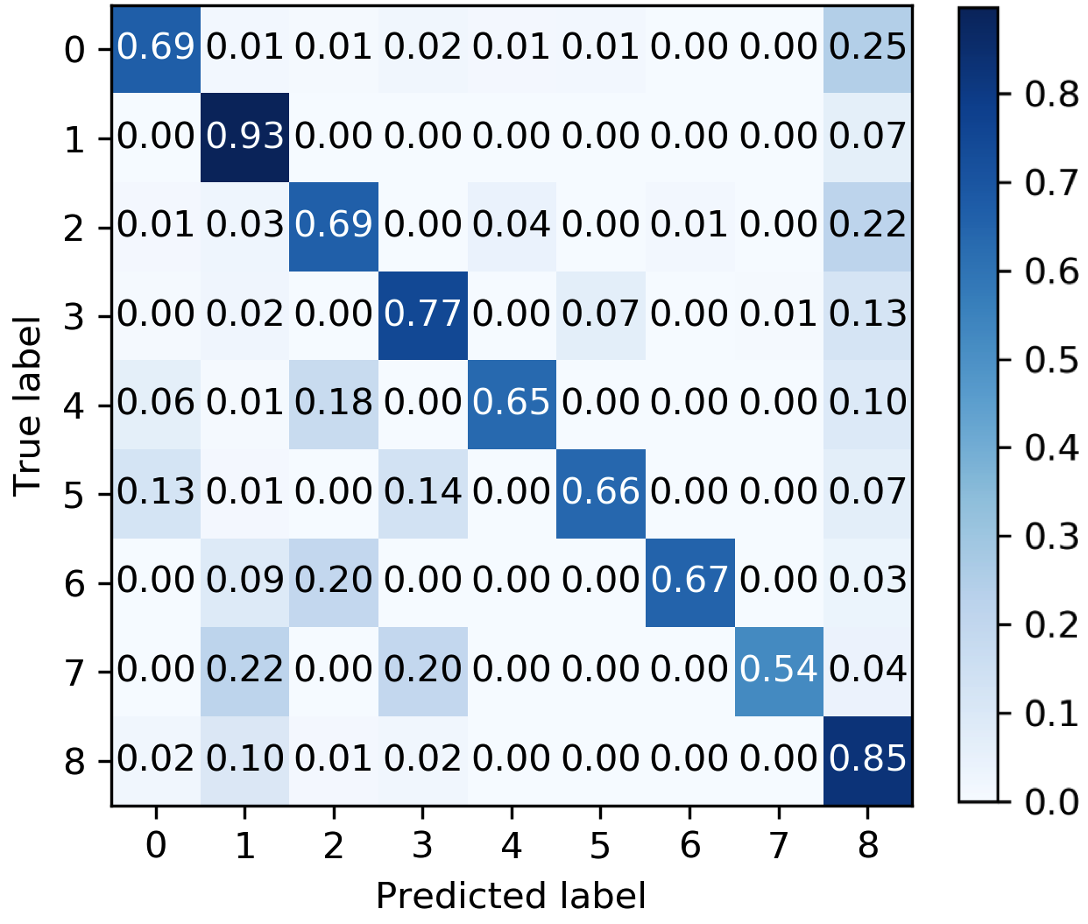}
	\end{minipage}
	}
 \hfill 	
  \subfloat[VGG16]{
	\begin{minipage}
	{
	   0.45\textwidth}
	   \centering
	   \includegraphics[width=1\textwidth]{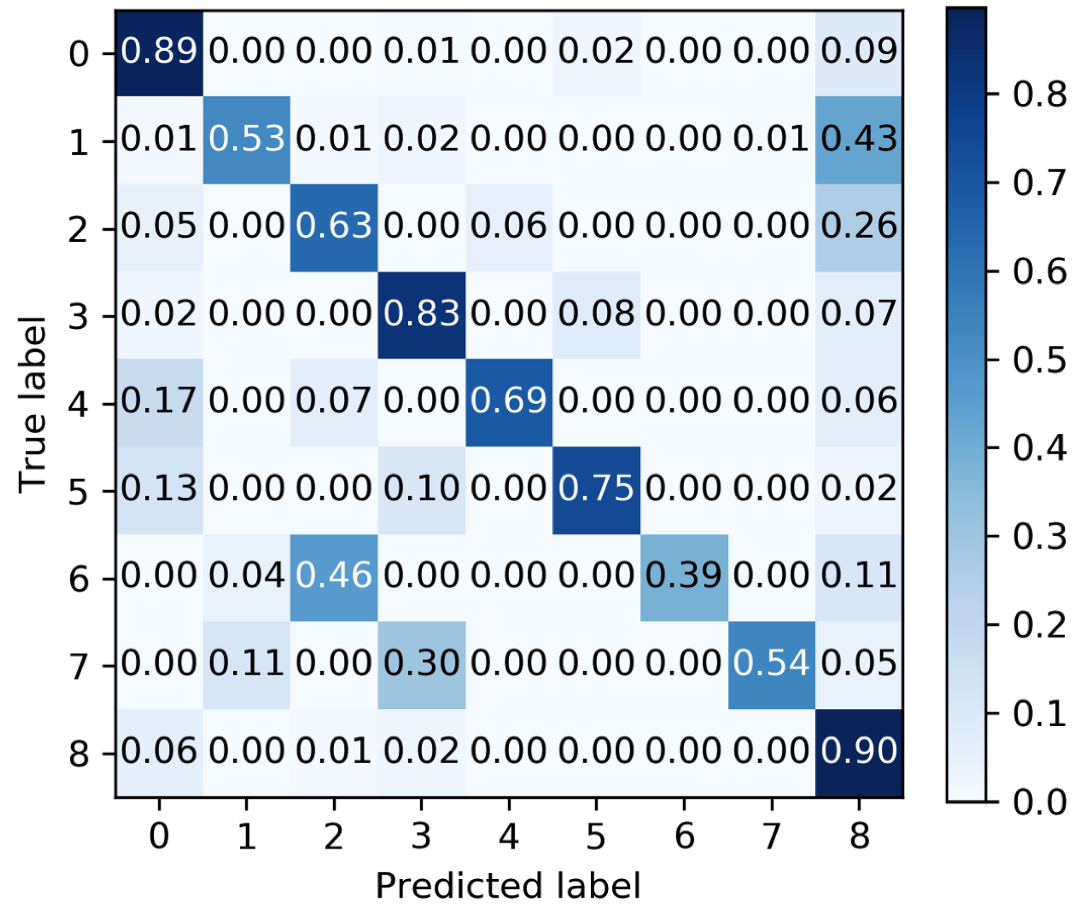}
	\end{minipage}
	}
 \hfill 
 \vfill
  \hfill 	
  \subfloat[VGG19]{
	\begin{minipage}
	{
	   0.45\textwidth}
	   \centering
	   \includegraphics[width=1\textwidth]{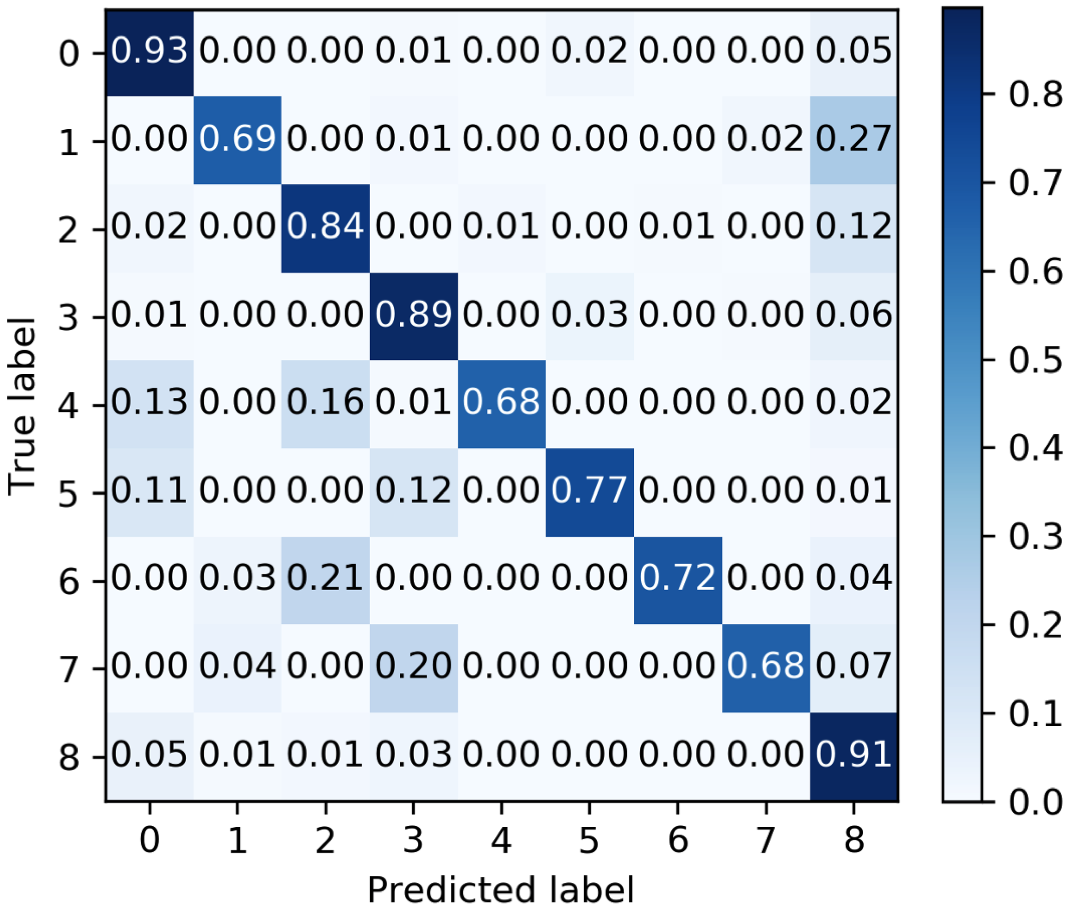}
	\end{minipage}
	}
 \hfill 	
  \subfloat[Xception]{
	\begin{minipage}
	{
	   0.45\textwidth}
	   \centering
	   \includegraphics[width=1\textwidth]{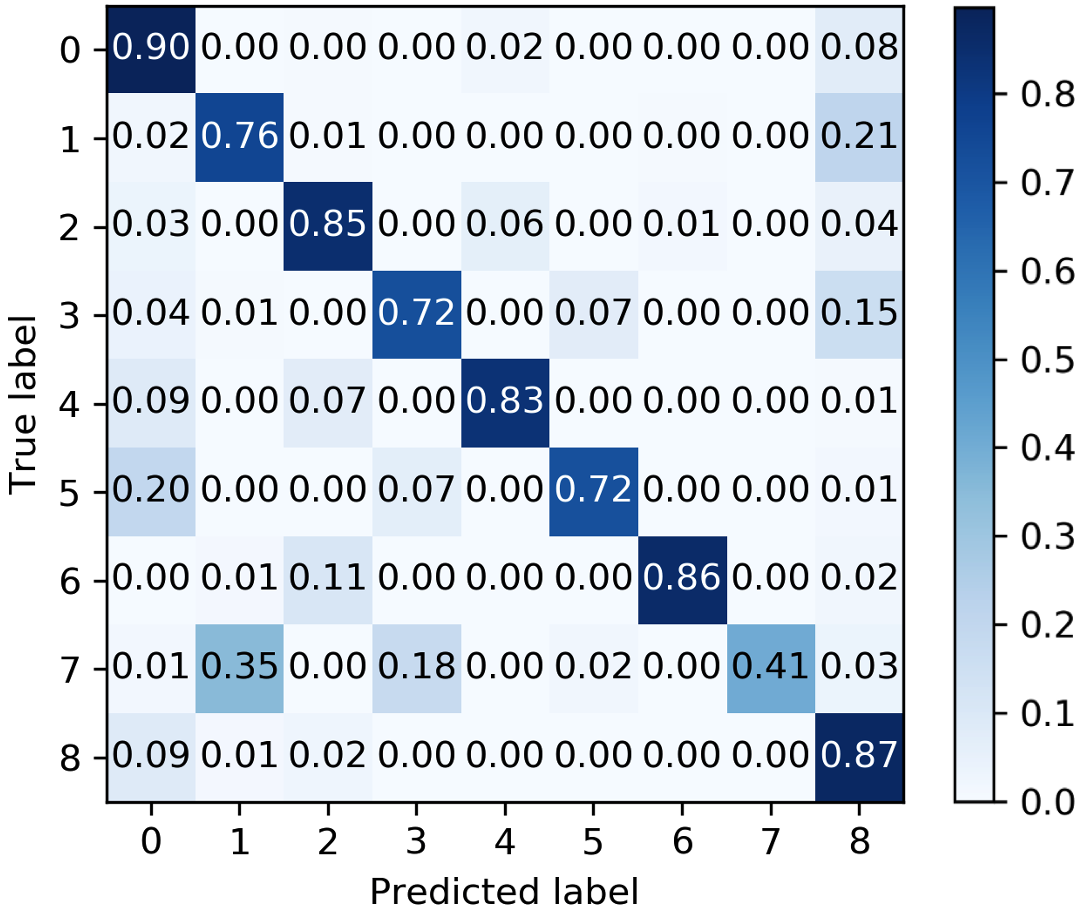}
	\end{minipage}
	}
 \hfill 
\caption{Class-wise normalized confusion matrices for 4 CNN architectures (i.e. EfficientNet, VGG16, VGG19, and Xception) used for attitude prediction using pilot attitude indicator gauge are presented.}
\label{fig:aai_cm_1}
\end{figure*}

Figure \ref{fig:pilot_ws_cm_1} presents class-wise normalized confusion matrices of four models trained on pilot windshield datasets. Figure \ref{fig:co_pilot_ws_cm_1} presents class-wise normalized confusion matrices of four models trained on co-pilot windshield datasets. The x-axis on for confusion matrices presents the predicted attitude class while the y-axis is reserved for the true class. The diagonal values show the correct classification rates for nine different attitude classes. The higher numbers on the diagonal of a confusion matrix reflect a better model. Figure \ref{fig:pilot_efis_cm_1} present class-wise normalized confusion matrices of four models trained on pilot EFIS display dataset. Figure \ref{fig:co_pilot_efis_cm_1} presents class-wise normalized confusion matrices of four models trained on co-pilot EFIS display datasets. Figure \ref{fig:aai_cm_1} presents class-wise normalized confusion matrices of four models trained on artificial attitude indicator gauge. 

\begin{figure*}[htb]
\centering
 \hfill 
  \subfloat[Excluding predictions of AAI models (i.e., 16 models)]{
	\begin{minipage}
	{
	   0.48\textwidth}
	   \centering
	   \includegraphics[width=1\textwidth]{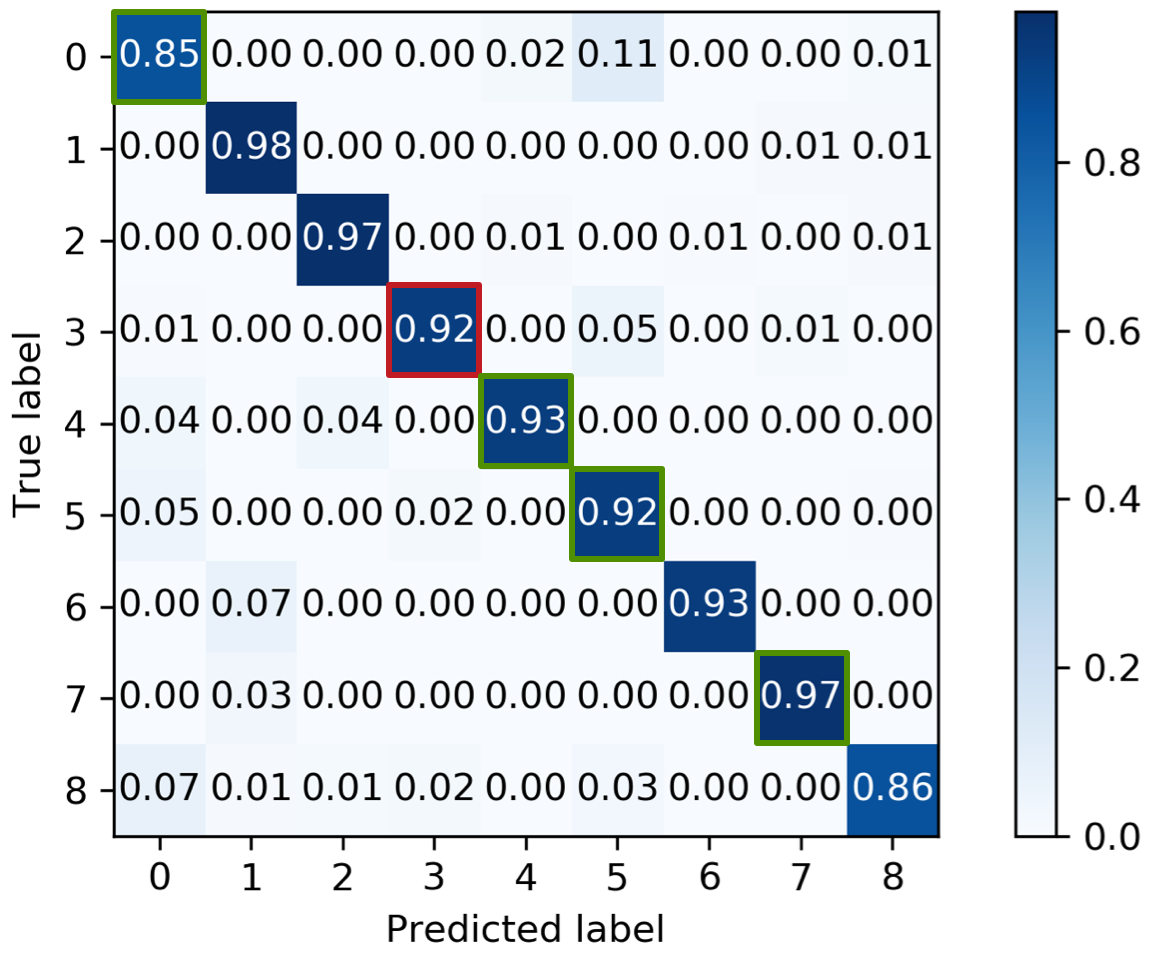}
	\end{minipage}
	}
 \hfill 	
  \subfloat[Considering predictions of all 20 models]{
	\begin{minipage}
	{
	   0.48\textwidth}
	   \centering
	   \includegraphics[width=1\textwidth]{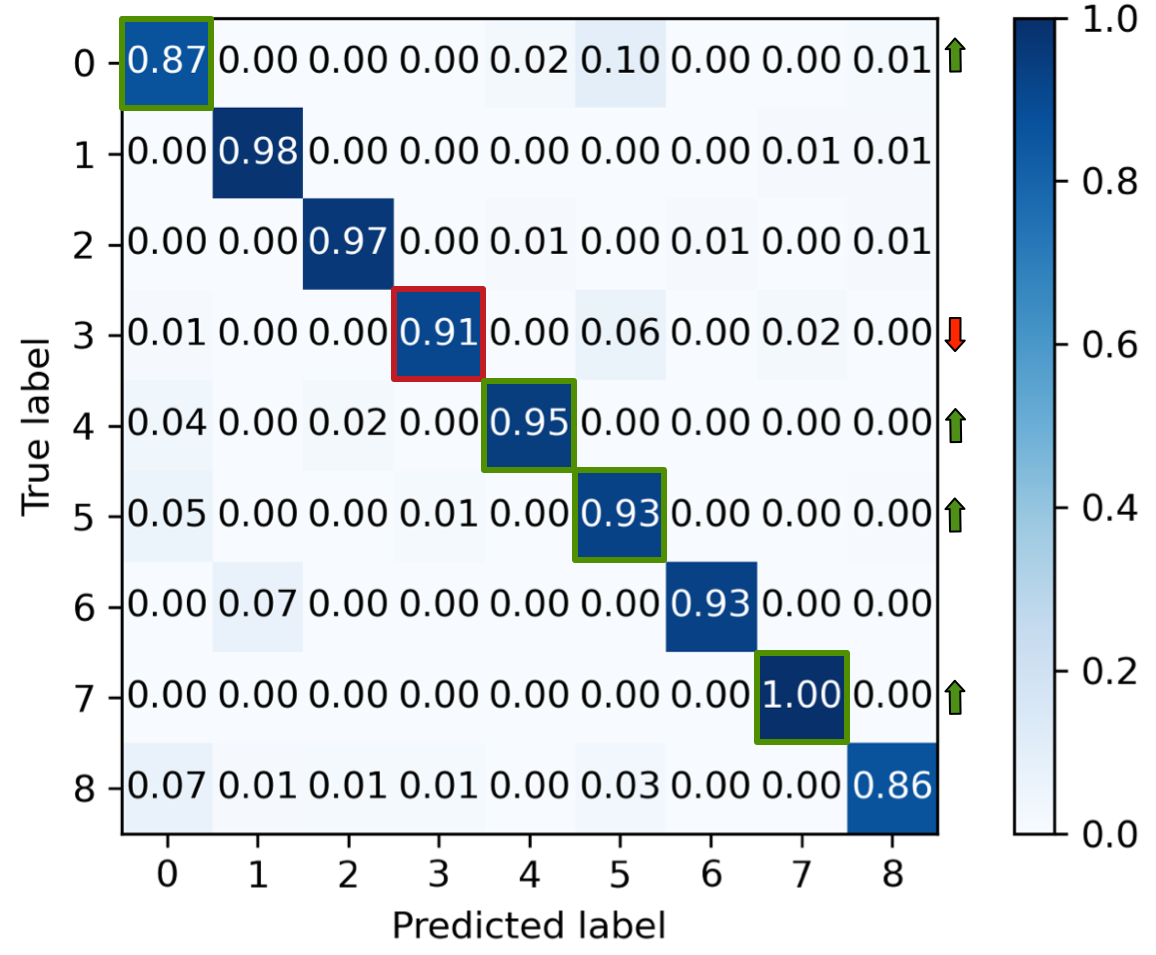}
	\end{minipage}
	}
 \hfill 
\caption{Class-wise normalized confusion matrices of ensemble attitude prediction are presented. We used the majority voting strategy to combine different model that are trained on five different camera views. (a) Ensemble confusion matrix based on 16 models trained on pilot windshield, co-pilot windshield, pilot EFIS and co-pilot EFIS views. (b) Ensemble confusion matrix for all 20 models. The ensemble approach obtained better accuracy at the individual class level and enhanced overall attitude predictive accuracy.}
\label{fig:Ensemble_CMs}
\end{figure*}

Figure \ref{fig:Ensemble_CMs} presents the normalized confusion matrices for both ensemble approaches. First, we considered 16 models trained on pilot and co-pilot windshield and pilot and co-pilot EFIS datasets in the first ensemble approach. While in the second, we considered all 20 models trained on datasets. In Figure \ref{fig:Ensemble_CMs}, we use green and red rectangles to show the positive (an increase in predicting accuracy) and negative (a decrease in the predictive accuracy) effects of the class-wise prediction in ensemble confusion matrices. We observed that increasing the number of models in the ensemble improved the prediction accuracy. It is also evident (red and green arrows) from Figure \ref{fig:Ensemble_CMs} that an increased number of models in the ensemble result in an increase in prediction accuracy for most classes. It is important to highlight that the ensemble-based approach obtained consistent results on all nine attitude classes. Whereas the models trained on specific camera views have comparatively lower classification accuracy values on classes 6 and 7.

\section{Conclusion}

We showed that ensemble AI models could reliably predict the rotorcraft attitude using data from ordinary cameras mounted in the cockpit. Previously, we only used a single camera view, which resulted in higher error rates and low reliability. Our approach overcame several common problems, i.e., obscure or invisible horizons during night flights, low visibility during precipitation or during fog/low clouds. We proposed two ensemble techniques that combine various trained CNNs models using a majority voting strategy. The ensemble approaches combine the visual information of all five camera views before making the final attitude predictions and obtain higher average class-level attitude predictive accuracies. Our results establish the potential of AI-based technologies for various rotorcraft safety applications.

\textbf{Authors' Contacts:}
\begin{itemize}
    \item Hikmat Khan:khanhi83@students.rowan.edu 

    \item Dr. Ghulam Rasool: Ghulam.Rasool@moffitt.org
    
    \item Dr. Nidhal C. Bouaynaya: bouaynaya@rowan.edu
    
    \item Charles C. Johnson: Charles.C.Johnson@faa.gov
\end{itemize}

\section{Acknowledgments}
This work was supported by the Federal Aviation
Administration (FAA) Cooperative Agreement Number 16-G-015 and NSF Awards OAC-2008690 and DUE-1610911. Ghulam Rasool was also partly supported by NSF Award OAC-2008690.
This publication was in part supported by a subaward from Rutgers University, Center for Advanced Infrastructure \& Transportation, under Grant no. 69A3551847102 from the U.S. Department of Transportation, Office of the Assistant Secretary for Research and Technology (OST-R).

\clearpage

\section{Authors Biography} 

\textbf{Hikmat Khan} is currently a PhD student at Rowan University. He is a research fellow supporting the Federal Aviation Administration (FAA) via a research grant/cooperative agreement by evaluating the feasibility of applying deep learning approaches to increase safety within the rotorcraft industry. His research interests include deep learning, continual learning, few-shot learning and optimization. 
\\

\textbf{Nidhal C. Bouaynaya} received her Ph.D. degree in Electrical and Computer Engineering and M.S. degree in Pure Mathematics from The University of Illinois at Chicago, in 2007. From 2007-2013, she was an Assistant then Associate Professor with the Department of Systems Engineering at the University of Arkansas at Little Rock. In Fall 2013, she joined the Department of Electrical and Computer Engineering at Rowan University, where she is currently a Professor and the Associate Dean for Research and Graduate Studies. Dr. Bouaynaya co-authored more than 100 refereed journal articles, book chapters and conference proceedings. She won numerous Best Paper Awards, the most recent was at the 2019 \emph{IEEE International Workshop on Machine Learning for Signal Processing}.  She is also the winner of the Top algorithm at the 2016 Multinomial Brain Tumor Segmentation Challenge (BRATS). Her research interests are in Big Data Analytics, Machine Learning, Artificial Intelligence and Mathematical Optimization. In 2017, she Co-founded and is Chief Executive Officer (CEO) of MRIMATH, LLC, a start-up company that uses artificial intelligence to improve patient oncology outcome and treatment response.
\\

\textbf{Ghulam Rasool} is an Assistant Professor of Electrical and Computer Engineering at Rowan University. He received a BS. in Mechanical Engineering from the National University of Sciences and Technology (NUST), Pakistan, in 2000, an M.S. in Computer Engineering from the Center for Advanced Studies in Engineering (CASE), Pakistan, in 2010, and the Ph.D. in Systems Engineering from the University of Arkansas at Little Rock in 2014. He was a postdoctoral fellow with the Rehabilitation Institute of Chicago and Northwestern University from 2014 to 2016. He joined Rowan University as an adjunct professor and later as a lecturer in the year 2018. Currently, he is the co-director of the Rowan AI Lab. His current research focuses on machine learning, artificial intelligence, data analytics, signal, image, and video processing. His research is funded by National Science Foundation (NSF), U.S. Department of Education, U.S. Department of Transportation (through the University Transportation Center (UTC), Rutgers University), Federal Aviation Administration (FAA), New Jersey Health Foundation (NJHF), and Lockheed Martin, Inc. His recent work on Bayesian machine learning won the Best Student Award at the 2019 IEEE Machine Learning for Signal Processing Workshop.
\\

\textbf{Charles C. Johnson} works as a research engineer, program manager, and technical expert for the Aviation Research Division at the FAA William J. Hughes Technical Center in Atlantic City, NJ. During his 10+ year-career with the FAA, he has led several rotorcraft and unmanned aircraft systems (UAS) research and development simulation/flight test activities that seek to improve aviation safety. Cliff is qualified on the ScanEagle UAS platform. He holds a Bachelor’s degree in Mechanical Engineering from Rowan University. He also holds a private pilot’s license (single engine land/fixed-wing) and is pursuing his instrument, commercial, and helicopter add-on ratings.
\\

\textbf{Tyler Travis} serves as a Research Analyst at the Federal Aviation Administration’s (FAA) William J. Hughes Technical Center in Atlantic City, NJ. She supports several research and development activities that seek to improve aviation safety. During her 5+ years at the FAA Technical Center, Tyler has worked on several UAS and Rotorcraft Human-In-The Loop simulations involving new technologies and procedural changes impacting the National Airspace System.  Prior to joining the FAA, Tyler completed her B.S. in Business Administration with a concentration in Management Information Systems at Drexel University in September 2012.
\\

\textbf{Lacey Thompson} works as an Operations Research Analyst for the Unmanned Aircraft Systems (UAS) Engineering Branch at the Federal Aviation Administration’s (FAA) William J. Hughes Technical Center in Atlantic City, NJ. During her 7.5 years with the FAA, Lacey has managed several UAS Human-In-the-Loop (HITL) simulations. In addition, she serves as an Aviation Science, Technology, and Mathematics (AvSTEM) Ambassador, creating the curriculum for the first ever module on UAS for the Aviation Monthly Mentoring Program. Lacey holds a Bachelor’s degree in Physics from Northwestern State University of Louisiana and a Master’s degree in Aeronautics from Embry-Riddle Aeronautical University Worldwide.
\\
\bibliographystyle{unsrt}
\bibliography{main.bib}
\end{document}